\documentclass[sigconf]{acmart}
\acmSubmissionID{5698}

\usepackage[utf8]{inputenc} % allow utf-8 input
\usepackage[T1]{fontenc}    % use 8-bit T1 fonts
\usepackage{hyperref}       % hyperlinks
\usepackage{url}            % simple URL typesetting
\usepackage{amsfonts}       % blackboard math symbols
\usepackage{microtype}      % microtypography
\usepackage{wrapfig}
\usepackage{multirow}
\usepackage{float}
\usepackage{math_commands}
\usepackage{subfigure}
\usepackage{algorithm} %format of the algorithm
\usepackage{algorithmic} %format of the algorithm
\usepackage{amsthm}
\usepackage{enumitem}
\usepackage{bm}

\theoremstyle{definition}

\setlist{nolistsep}
\usepackage{xspace}
\mathchardef\mhyphen="2D

\newcommand{\method}{KGCF\xspace}
\newcommand{\best}[1]{\textbf{#1}}
\newcommand{\xhdr}[1]{\noindent{{\bf #1.}}}
\newcommand{\edge}[1]{$\langle$\textit{#1}$\rangle$}
\newcommand{\func}[1]{\ensuremath{\mathrm{#1}}}
\newcommand{\rank}[0]{\func{rank}}
\newcommand{\set}[1]{\ensuremath{\mathcal{#1}}}

\AtBeginDocument{%
  \providecommand\BibTeX{{%
    \normalfont B\kern-0.5em{\scshape i\kern-0.25em b}\kern-0.8em\TeX}}}

\setcopyright{acmcopyright}
\copyrightyear{2023} 
\acmYear{2023} 
\setcopyright{acmlicensed}\acmConference[WWW '23]{Proceedings of the ACM Web Conference 2023}{April 30-May 4, 2023}{Austin, TX, USA}
\acmBooktitle{Proceedings of the ACM Web Conference 2023 (WWW '23), April 30-May 4, 2023, Austin, TX, USA}
\acmPrice{15.00}
\acmDOI{10.1145/3543507.3583401}
\acmISBN{978-1-4503-9416-1/23/04}

\begin{document}

%%
%% The "title" command has an optional parameter,
%% allowing the author to define a "short title" to be used in page headers.
\title{Knowledge Graph Completion with Counterfactual Augmentation}

%%
%% The "author" command and its associated commands are used to define
%% the authors and their affiliations.
%% Of note is the shared affiliation of the first two authors, and the
%% "authornote" and "authornotemark" commands
%% used to denote shared contribution to the research.

% \author{Ben Trovato}
% \authornote{Both authors contributed equally to this research.}
% \email{trovato@corporation.com}
% \orcid{1234-5678-9012}
% \author{G.K.M. Tobin}
% \authornotemark[1]
% \email{webmaster@marysville-ohio.com}
% \affiliation{%
%   \institution{Institute for Clarity in Documentation}
%   \streetaddress{P.O. Box 1212}
%   \city{Dublin}
%   \state{Ohio}
%   \country{USA}
%   \postcode{43017-6221}
% }

\author{Heng Chang}
\email{changh17@tsinghua.org.cn}
\orcid{0000-0002-4978-8041}
\affiliation{%
  \institution{Tsinghua University}
}

\author{Jie Cai}
\orcid{0000-0003-2191-8585}
\email{caij20@mails.tsinghua.edu.cn}
\affiliation{%
  \institution{Tsinghua University}
}

\author{Jia Li}
\orcid{0000-0002-6362-4385}
\authornote{Corresponding author.}
\email{jialee@ust.hk}
\affiliation{%
\institution{Hong Kong University of Science and Technology (Guangzhou)}
}

%%
%% By default, the full list of authors will be used in the page
%% headers. Often, this list is too long, and will overlap
%% other information printed in the page headers. This command allows
%% the author to define a more concise list
%% of authors' names for this purpose.
% \renewcommand{\shortauthors}{Chang, et al.}

%%
%% The abstract is a short summary of the work to be presented in the
%% article.
\begin{abstract}
  Graph Neural Networks (GNNs) have demonstrated great success in Knowledge Graph Completion (KGC) by modeling how entities and relations interact in recent years. However, most of them are designed to learn from the observed graph structure, which appears to have imbalanced relation distribution during the training stage. Motivated by the causal relationship among the entities on a knowledge graph, we explore this defect through a counterfactual question: ``\emph{would the relation still exist if the neighborhood of entities became different from observation?}''.
  With a carefully designed instantiation of a causal model on the knowledge graph, we generate the counterfactual relations to answer the question by regarding the representations of entity pair given relation as context, structural information of relation-aware neighborhood as treatment, and validity of the composed triplet as the outcome.
  Furthermore, we incorporate the created counterfactual relations with the GNN-based framework on KGs to augment their learning of entity pair representations from both the observed and counterfactual relations.
  Experiments on benchmarks show that our proposed method outperforms existing methods on the task of KGC, achieving new state-of-the-art results. Moreover, we demonstrate that the proposed counterfactual relations-based augmentation also enhances the interpretability of the GNN-based framework through the path interpretations of predictions.
\end{abstract}

%%
%% The code below is generated by the tool at http://dl.acm.org/ccs.cfm.
%% Please copy and paste the code instead of the example below.
%%


%%
%% Keywords. The author(s) should pick words that accurately describe
%% the work being presented. Separate the keywords with commas.

%%
%% This command processes the author and affiliation and title
%% information and builds the first part of the formatted document.
\maketitle

\section{Introduction}\label{sec:intro}
Knowledge can be encoded in knowledge graphs (KGs), which store structured information of real-world entities as nodes and relations as edges. Real-world KGs are
usually noisy and incomplete.
Because of the nature of incompleteness of KGs, knowledge graph completion (KGC) is designed to predict missing relations~\cite{bordes2013translating,galarraga2015fast,ho2018rule}.

\begin{figure}[tb]
\centering
\subfigure[Train split of FB237]{
\includegraphics [width=0.485\columnwidth]{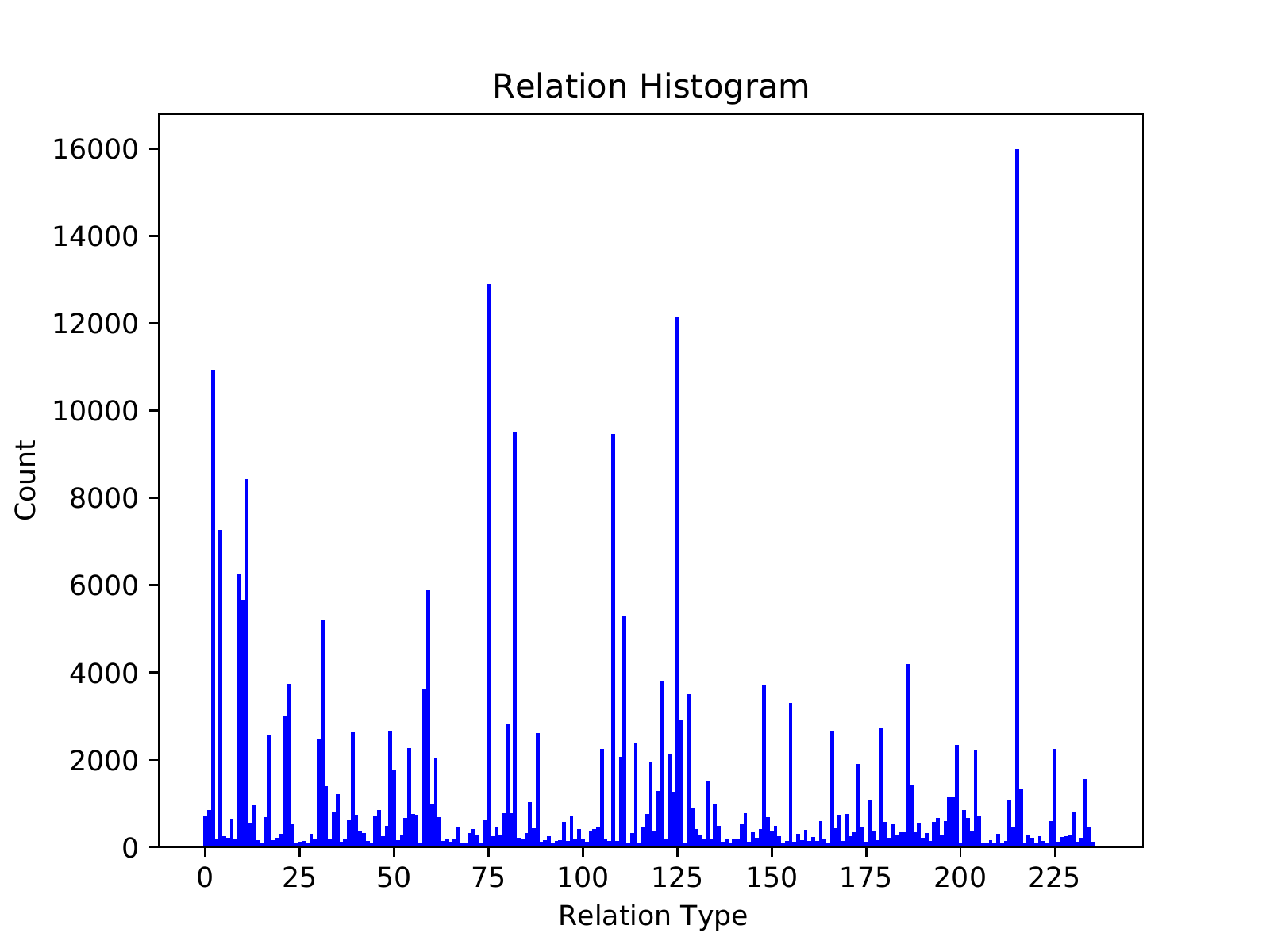}}
\subfigure[Train split of WN18RR]{
\includegraphics [width=0.485\columnwidth]{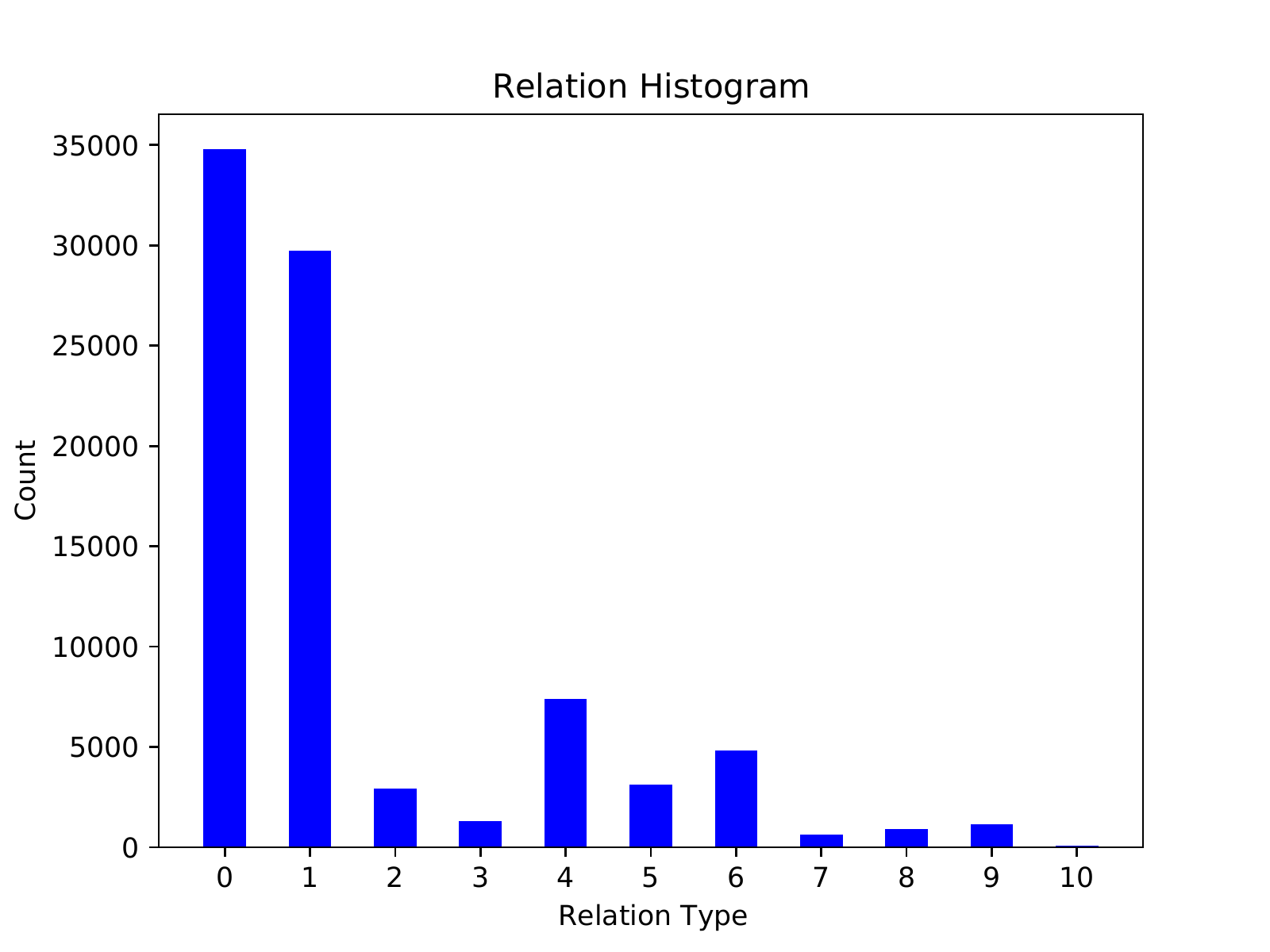}}
\vspace{-2mm}
\caption{Relation type distribution from the train and test split on dataset FB237 and WN18RR, respectively.}
\Description{Relation type distribution from the train and test split on dataset FB237 and WN18RR, respectively, which is obviously imbalanced.}
\vspace{-6mm}
\label{fig:rel-distribution}
\end{figure}

GNN-based KGC models have shown effects on the KGC task and attracted tremendous attention in recent years~\cite{ji2021survey}. Many popular GNNs follow an iterative message passing then aggregate scheme to adaptively learn the representations of nodes in graph~\cite{zhang2020deep,chang2020restricted,chang2022adversarial,guan2021autogl}. When learning on a homogeneous graph, the representation of the center node is updated by aggregating the message gathered from the neighbors of the node while preserving the local structure around the node during each iteration, 
While a homogeneous graph could be viewed as a special version of KGs, 
GNNs are not able to be directly employed on KGs and are usually modified to model the interactions between entities and multiple relations. However, as shown in Figure~\ref{fig:rel-distribution}, the distribution of relation types is usually imbalanced on KGs. Most of the current GNN-based approaches fail to capture this information and could not generalize well. This failure calls for a better learning regime on KGC.

\begin{figure}[tb]
\centering
\includegraphics [width=0.98\columnwidth]{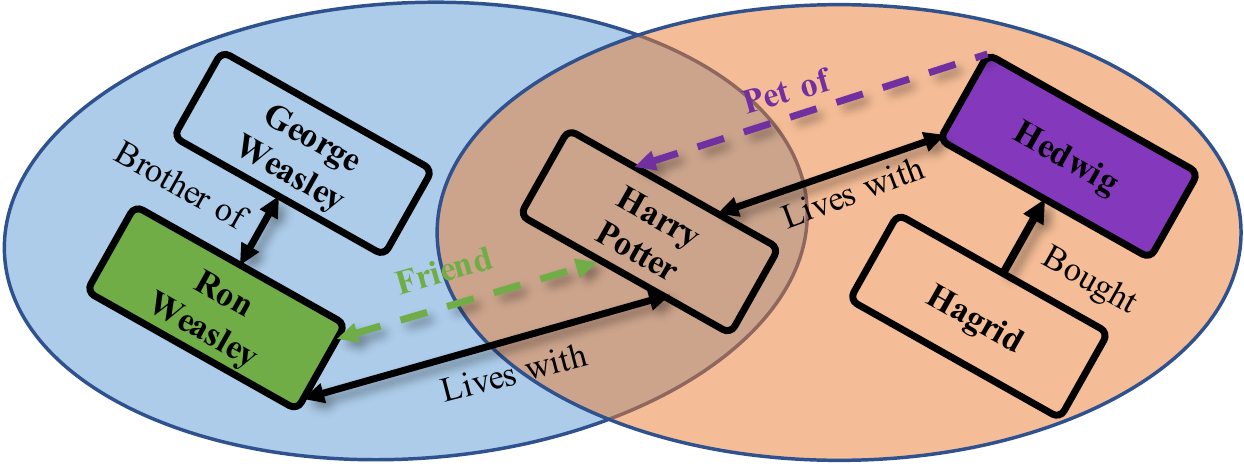}
\vspace{-2ex}
\caption{Example of why simply considering neighborhood information for causal learning from a homogeneous graph view would fail on KGs.
}
\Description{Example of why simply considering neighborhood information for causal learning from a homogeneous graph view would fail on KGs. Let the blue and orange ovals be two detected communities by treating the KG as a homogeneous graph. Hedwig (an owl highlighted as purple) and Ron Weasley (a person highlighted as green) have the same neighborhood and would be assigned with the same treatment in the causal model. While the missing two relations are different, this definition of treatment is not beneficial to representation learning and prediction.}
\vspace{-6mm}
\label{fig:kg-exp}
\end{figure}

In this work, we propose to alleviate this failure with data augmentation on KGs, especially from the perspective of causal inference by answering the following counterfactual question:

\fbox{\begin{minipage}{23.6em}
Would the relation still exist if the neighborhood of entities became different from observation?
\end{minipage}}

If a GNN-based KGC model learns this causal relationship via answering the counterfactual question above, such knowledge will help to discover causal effects on relation types and improve the accuracy of prediction as well as the generalization ability~\cite{zhao2022learning}. Note that the causality here falls into the scope of causal inference models, where we focus on taking advantage of existing basic yet effective causal models to enhance the task of KGC. This aligns with a line of research works~\cite{johansson2016learning,alaa2019validating,zhao2022learning}, and is not related to the explicit semantic relations in KGs.

A counterfactual question is often associated with three factors: context (as a data point), treatment (as a type of manipulation), and outcome~\cite{van2007causal,johansson2016learning,zhao2022learning}. Previous effort~\cite{zhao2022learning} proposes to use whether two nodes live in the same neighborhood to define such treatment on a homogeneous graph. However, we argue that this definition has flaws in KGs since it ignores the specific relation connected entities. Taking Figure~\ref{fig:kg-exp} as an example. 
Let the blue and orange ovals be two detected communities by treating the KG as a homogeneous graph.
Hedwig and Ron Weasley obviously share the same neighborhood within the ovals if we ignore the relation type on edges. Then the treatment definition from~\cite{zhao2022learning} would be confused and make no distinction between Ron Weasley, who is a person, and Hedwig, which is an owl. As a result, it is even harder to help predict the different relations "\textit{pet of}" and "\textit{friend}" that are associated with them and Harry Potter.
Therefore, it is non-trivial to directly migrate the treatment definition from a homogeneous graph to KGs due to the multi-relation nature.

Based on the idea of considering relation types with causal learning on KGs,
we propose an instantiation of causal models on KGs with a new definition of treatment for context.
By answering the counterfactual questions, we discover the counterfactual relations for all observed contexts and calculate their counterfactual treatments accordingly.
Then, we integrate the representation learning of a pair of entities given a query relation with both their factual and counterfactual treatments and propose a counterfactual relation augmented GNN-based framework (\method) for KGC. 
Specifically, since this counterfactual treatment is not available from the observed data (valid triplets on KGs), we match it with the nearest neighbor of an entity pair in hidden embedding space and use the neighbor's factual treatment as a substitution. 
To further address the imbalanced distribution of relation types, we first embed the entities to an embedding space that is associated with each relation type, then find the nearest neighbors within this embedding space. After obtaining both the factual and counterfactual views of training data, we consociate them into an encoder-decoder style framework to learn better representations and predict missing relations.

The contributions of this paper are summarized as follows:
\begin{itemize}[noitemsep,topsep=0pt,parsep=0pt,partopsep=0pt,leftmargin=*]
    \item Motivated by the need for data augmentation on KGs, we propose the first instantiation of the causal model for KGs via answering counterfactual questions and considering the relation types.
    \item We present \method that utilizes counterfactual relations to augment the representation learning on KGs with special consideration on the imbalanced relation distribution.
    \item Extensive experiments show the superiority of \method on KGC task. We further demonstrate that counterfactual relations also enhance the interpretation ability through path-based explanation of predictions. 
\end{itemize}

\section{Related Work}\label{sec:related-works}
\subsection{Knowledge Graph Completion (KGC)}
Popular research KGC models mainly fall into three categories~\cite{ji2021survey}: embedding-based methods, relation path inference, and rule-based reasoning. While the last two research directions focus on exploring the multi-step relationships, we focus on the recently preliminary direction of KGC in this paper: embedding-based KGC methods with GNNs and cover some of the recent progress in this line of research due to a large amount of literature.
% (2021 KDD) T-GAP: Learning to Walk across Time for Temporal Knowledge Graph Completion
\citet{jung2020t} propose a new GNN encoder, which can effectively capture the query-related information from the temporal KG and is interpretable in the reasoning process.
% (2022 WWW) Rethinking Graph Convolutional Networks in Knowledge Graph Completion
\citet{zhang2022rethinking} verify by experiments that the graph structure modeling in GCNs has no significant impact on the performance of the KGC model. Instead, they propose a simple and effective LTE-KGE framework by removing the GCN aggregation.
% (2022 AAAI) Exploring Relational Semantics for Inductive Knowledge Graph Completion
\citet{Wang_Zhou_Pan_Dong_Song_Sha_2022} propose a new model called CFAG, which uses a coarse-grained aggregator (CG-AGG) and a fine-grained generative adversarial net (FG-GAN) to resolve the emergence of new entities.
% (2022 ACL) Multilingual Knowledge Graph Completion with Self-Supervised Adaptive Graph Alignment
\citet{huang2022multilingual} propose a KG inference framework SS-AGA based on self-supervised graph alignment. Despite the progress of the above-mentioned models, most of them fail to consider the imbalanced distribution of relation types.

\subsection{Causal Inference}
Finding causal relationships between different variables from observational data is a long-standing scientific problem in the field of statistics and artificial intelligence~\cite{yao2020causal}.
Causal inference for GNNs is an emerging field of research, which can be divided into three categories: counterfactual learning, causal graph-guided representation learning, and causal discovery.
% (2021 ICML) Generative Causal Explanations for Graph Neural Networks
\citet{lin2021generative} use causal inference to improve the explainability of GNNs. They transform the explainability problem of reasoning decisions in the GNNs into a causal learning task, and then a causal explanation model is trained based on the objective function of Granger causality.
% (2022 KDD) Learning Causal Effects on Hypergraphs
\citet{ma2022learning} propose a new framework called causal inference under spillover effects, which solves the problem of individual treatment effect estimation with high-order interference on hypergraphs.
% (2022 ICML) Learning from Counterfactual Links for Link Prediction
\citet{zhao2021counterfactual} first propose to use causal inference to improve link prediction on homogeneous graphs, which is the most related one to this paper. They propose the concept of counterfactual link prediction (CFLP) and representations are learned from observed and counterfactual links which are used as augmented training data. In contrast, we focus on KGs in this paper, to which CFLP could not be trivially migrated as discussed in Sec.~\ref{sec:intro}.

\subsection{Data Augmentation on Knowledge Graph}
Graph Data Augmentation (GraphDA) is to find a mapping function that enriches or changes the information in a given graph~\cite{gu2020implicit,jin2021power,chang2021not,chang2021spectral}, which is a relatively underexplored area for KGs.
\citet{CHEN2021352} propose a rumor data augmentation method called graph embedding-based rumor data augmentation (GERDA) to solve the data imbalance of rumor detection, which uses KGs to simulate the rumor generation process from the perspective of knowledge.
\citet{tang2022positive} propose positive-unlabeled learning with adversarial data augmentation (PUDA). PUDA alleviates the problem of false negatives through positive-unlabeled learning and positive sample sparsity through adversarial data augmentation.
Different from them, we propose to use the causal model to generate counterfactual augmentations to facilitate representation learning on KGs.

\section{Preliminaries}\label{sec:pre}
\subsection{Knowledge Graphs (KGs)}
A knowledge graph (KG) is denoted as a set of triplets: 
\begin{equation}
    \notag \gG = \{(e_i,r_j,e_k)\} \\ \subset (\gE \times \gR \times \gE), 
\end{equation}
where $\gE$ and $\gR$ represent a set of entities (nodes) and relations (edges), respectively. $e_i$ and $r_j$ are the $i$-th entity and $j$-th relation, and the types of relations could be in great numbers within a KG. We also usually distinguish the entity pair $(e_i, e_k)$ with $(h_i, t_k)$, considering the direction of the two entities. 
Note that a homogeneous graph $\gG=(\gV, \gE)$ can be viewed as only one relation type for all edges versions of KGs. 
Throughout this paper, we use \textbf{bold} terms, $\mW$ or $\ve$, to denote matrix/vector representations for weights and entities, respectively. And we select \textit{italic} terms, $w_h$ or $\alpha$, to denote scalars.
We can also use a third-order binary tensor $\gA \in\{0,1\}^{|\gE|\times|\gR|\times|\gE|}$ to uniquely define a KG, which is also known as the adjacency tensor of $\mathcal{G}$. The $(i,j,k)$ entry $\gA_{ijk}=1$ if $(h_i, r_j, t_k)$ is valid or otherwise $\gA_{ijk}=0$. The $j$-th frontal slice of $\gA$ is the adjacency matrix of the $j$-th relation $r_j$, which we denote as $\gA_{r_j}$ for specification. To ease the notation, we also use $\gA_{(h, r, t)}$ to denote the entry of a given triplet $(h, r, t)$ in the adjacency tensor.

\subsection{GNN-Based framework for KGC}\label{sec:pre_gnnkgc}
GNN-Based framework dealing with KGC usually adopts an encoder-decoder style framework~\cite{schlichtkrull2018modeling}, where GNNs perform as the encoder and embedding-based score functions (e.g., TransE, DistMult, RotatE and ConvE) perform as the decoder. A KG's entities and relations are first represented as embeddings by the GNN encoder.
After having the embeddings from GNN (encoder) for entities and relations, KGC models usually simulate how entities and relations interact with the help of a score function (decoder) $s:\gE\times\gR\times\gE\rightarrow \mathbb{R}$. 
The decoder then predicts the entries in the adjacency tensors $\gA$ utilizing the derived representations. The prediction could be viewed as a recovery of the original graph structures since adjacency tensors and graph structures are in bijection. In this way, the score function-based decoder can detect missing relations in the original KG while completing the KG by recovering the graph structure. 
It is worth mentioning that in some GNN-based frameworks for KGC, such as NBFNet, the aforementioned score functions are implicitly included in the procedure of message passing, and a simple feed-forward neural network is employed as decoder on the learned entity pair representations to predict the missing links given the head/tail entity.

\begin{figure}[tb]
\centering
\includegraphics [width=0.95\columnwidth]{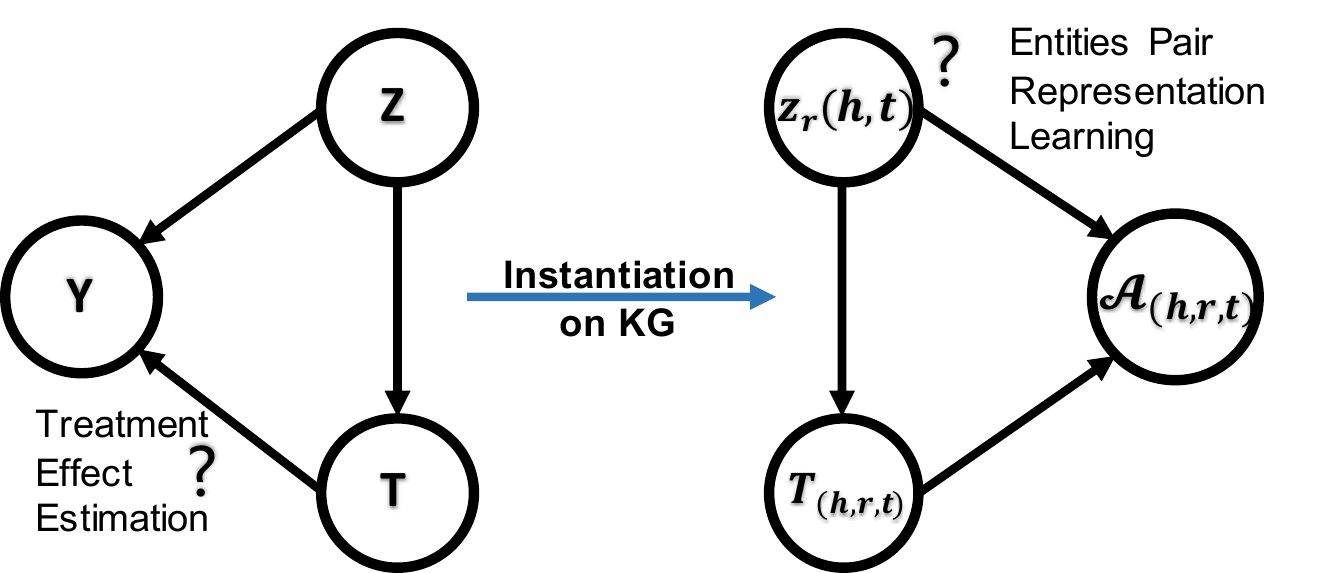}
\caption{Our proposed method improves the prediction of missing relations by leveraging a causal model on the KG. Left: Illustration of a typical causal DAG model.
Right: Our proposed KGC learning with the causal model. 
}
\Description{Our proposed method improves the prediction of missing relations by leveraging a causal model on the KG. Left: Illustration of a typical causal DAG model.
Given Z and outcome observations, the causal model aims to find the treatment effect of T on Y. 
Right: Our proposed KGC learning with the causal model. 
We instantiate the typical causal model on the knowledge graph by leveraging the estimated $\text{ITE}(\gA_{(h, r, t)}|\gT_{(h,r,t)})$ to improve the learning of pair representation $\vz_{r}(h,t)$, where $\vz_{r}(h,t)$ is the embedding of head-tail pair $(h,t)$ with a given relation $r$.}
\vspace{-6mm}
\label{fig:dag}
\end{figure}

\section{Proposed Method}\label{sec:method}
\subsection{Incorporating GNN-based framework for KGC with Causal Model}\label{sec:causalmodel}
Previous works on GNN-based framework for KGC~\cite{ji2021survey} have shown that message passing and aggregating scheme is able to generate more structure-enriched representations. 
However, recalling the example in Figure~\ref{fig:rel-distribution}, the distribution of relations appears to be quite imbalanced in KGs. This imbalanced distribution might lead the learning of GNNs to be trapped into the dominant relation types and fail to generalize to other relations during inference. Nevertheless, with the help of recent progress of data augmentation on graph learning~\cite{zhu2021graph,li2022semi,ding2022data}, 
we propose to enhance the generalization ability of current GNN-based learning frameworks on KGC from the perspective of a causal model.

Counterfactual causal inference seeks to determine the causal relationship between treatment and outcome by posing counterfactual questions. An illustration of the counterfactual question is ``would the outcome be different if the treatment was different?''~\cite{morgan2015counterfactuals}. 
In order to address the counterfactual question, researchers would next develop a causal model and apply causal inference accordingly.
For example, the left subfigure from Figure~\ref{fig:dag} is a typical triangle causal-directed acyclic graph (DAG) model. In this example, we denote the context (confounder) as $\mZ$, the treatment as $\mT$, and the outcome as $\mY$.
Once three of them are decided, the counterfactual inference methods could determine the effect of treatment $\mT$ on the outcome $\mY$ given context $\mZ$. Two statistics, \emph{individual treatment effect} (ITE) and its expectation \emph{averaged treatment effect} (ATE)~\cite{van2007causal,weiss2015machine,zhao2022learning}, are usually employed to measure this effect.
Given a binary treatment variable $T = \{0, 1\}$, we denote $g(\mathbf{z}, T)$ as the outcome of $\mathbf{z}$, then
\begin{equation}
    \notag \text{ITE}(\vz) = g(\vz, 1) - g(\vz, 0), \;\;\; \text{ATE} = \mathbb{E}_{\vz\sim \mZ}~\text{ITE}(\vz).
\end{equation}
Bigger ATE/LTE indicates a stronger causal relationship between treatment and outcome. 
Recent effort~\cite{zhao2022learning} proposes to develop a causal model to enhance the representation learning on graphs. 
Through proper definitions of context, treatments, and their corresponding outcomes on graphs,
promising performance on plain link prediction problems for homogeneous graphs is obtained.

On the contrary, in this work, we focus on a more comprehensive version of graphs, the KG, and focus on the KGC task. Given a pair of head entities and relation
$(h, r, ?)$, our model needs to predict the existence of a tail relation $t$. Hence, we aim to learn an effective representation $\vz_r(h, t)$ of  head-tail pair $(h,t)$ given a query relation $r$. 
As shown in the right subfigure from Figure~\ref{fig:dag}, we define the context as the entity pair representation $\vz_{r}(h, t)$, and the outcome in the adjacency tensor $\gA_{(h,r,t)}$ is the validity of triplet $(h, r, t)$. 
Therefore, our causal DAG contains the following nodes:
\begin{itemize}[leftmargin=*]
    \item $\vz$ (Confounder; unobservable): latent entity pair representations;
    \item $\gT$ (Treatment; observable): KG structural information (e.g., whether two entities belong to the same neighborhood);
    \item $\gA$ (Outcome; observable): triplet existence.
\end{itemize}
And the edges in our DAG are: $\vz$ -> $\gT$, $\vz$ -> $\gA$ and $\gT$ -> $\gA$.

In this way, we instantiate the typical causal model on the KG by leveraging the estimated $\text{ITE}(\gA_{(h, r, t)}|\gT_{(h,r,t)})$ to improve the learning of $\vz_{r}(h,t)$.
Here, the objective is different from classic causal inference, as well as the definition of the plain link prediction in~\cite{zhao2022learning}, considering the multiple types of relations.
More specifically, for each pair of entities $(h, t)$ given a query relation $r$, the corresponding ITE can be estimated by
\begin{equation}
\label{eq:ite}
    \text{ITE}_{(h,t)} = g(\vz_{r}(h, t), 1) - g(\vz_{r}(h, t), 0).
\end{equation}

In the next, we will incorporate the information of ITE from the counterfactual relations to augment the KG and improve the learning of $\vz_{r}(h, t)$, accordingly.
Both the entity pair representation $\vz_{r}(h, t)$ and the outcome $\gA_{(h,r,t)}$ need to consider the presence of multiple types of relations. Furthermore, the imbalanced distribution also makes the design of incorporation to the GNN-based embedding framework on KGC difficult.

As a preparation for the next step of incorporation, we denote the observed adjacency tensor by $\gA^{F}$ as \emph{factual} outcomes. In other words, $\gA^{F}$ is constructed with all triplets from the training set.
We also denote the unobserved tensor of the counterfactual relations by $\gA^{CF}$ when the treatment is different from the \emph{counterfactual} outcomes.
For the definition of treatment, we denote $\gT^{F} \in\{0,1\}^{|\gE|\times|\gR|\times|\gE|}$ as the binary factual treatment tensor, where $\gT^{F}_{(h, r, t)}$ indicates the treatment of the entity pair $(h, t)$ given a query relation $r$. 
Then we denote $\gT^{CF}$ as the counterfactual treatment matrix where $\gT^{CF}_{(h, r, t)} = 1 - \gT^{F}_{(h, r, t)}$. We are interested in two aspects: \textbf{1)} Estimating the counterfactual outcomes $\gA^{CF}$ \textbf{2)} learning from both factual (observed data) outcomes $\gA^{F}$ and counterfactual (augmented data) outcomes $\gA^{CF}$ to enhance the ability of GNN-based framework on KGC.

\subsection{Treatment Variable on KG}\label{sec:TV}
In this section, we describe how we design to acquire the factual treatment variable $\gT^{F}$.
As in many GNN-based embedding frameworks for KGC, we assume the entities that share similar neighborhoods would tend to share similar relations as well.
Therefore, finding the individual treatment for a pair of entities from their neighborhood under a special consideration to relation type is intuitively effective for downstream tasks.
Meanwhile, as pointed out by recent works on homogeneous graphs~\cite{cai2021line,tang2022rethinking,zhao2022learning}, models may fail to identify more global but also important factors purely from the association between local structural information and link presence.
Therefore, in this work, we use both local and global structural roles of each entity pair as its treatment. 

In KGs, since the relations are divided into multiple types, we need to define treatment based on the adjacency tensor $\gA_{r}$ w.r.t the query relation $r$, rather than simply dealing with a collapsed adjacency matrix as the regular practice in GNN-based approaches~\cite{schlichtkrull2018modeling,shang2019end,vashishth2019composition}.
Note that though in this work, we define the treatment from the structural information w.r.t the view of each relation $r_j$ on KGs for illustration, the causal models shown in \ref{fig:dag} do not limit the treatment to be structural roles on the graph. The treatment variable $\gT^{F}_{(h,r,t)}$ can be defined from any property of triplet $(h, r, t)$.

Then for each adjacency tensor in $\gA_{r_j}$, we define the binary treatment variable according to whether the two entities in a pair belong to the same community/neighborhood from a global perspective and calculate them separately. In this way, the relation types are explicitly considered in our treatment definition.
We denote $c: \mathcal{V} \rightarrow \mathbb{N}$ as any graph clustering or community detection approach that assigns each entity with an index of the community/cluster/neighborhood that the entity belongs to. The treatment tensor $\gT$ is defined as $\gT^{F}_{(h,r,t)} = 1$ if $c(h) = c(t)$ on adjacency tensor $\gA_{r}$, and $\gT^{F}_{(h,r,t)} = 0$ otherwise.
For the choice of $c$, without the loss of generality, an unsupervised approach K-core~\citep{malliaros2020core} that is widely used for clustering graph structure is utilized in our work as an example.
K-core is strongly related to the concept of graph degeneracy, which has a long history in graph theory.
Given a graph $\gG$ and an integer $k$, a K-core of a graph $\gG$ is a maximal connected subgraph of $\gG$ in which all vertices have degrees at least $k$. Equivalently, it is the connected components that are left after all vertices of degree less than $k$ have been removed.

The reason that we choose the K-core algorithm falls into three aspects: \textbf{1)} The core decomposition-based approach can be used to quantify node importance in many different domains efficiently and effectively. Precisely, the K-core algorithm for a graph $\gG$ can be computed in linear time w.r.t the number of edges $\gE$ of $\gG$, which is of great importance considering the massive entities in a KG. 
\textbf{2)} By assigning to each graph node an integer number $k$ (the core number), K-core could capture how well the given node is connected w.r.t its neighbors, which could also reflect the first-order proximity on graph~\cite{WWW2015Line}, which is a desired structural property for neighborhood. 
\textbf{3)} As one of the generally used clustering methods that consider the global graph structural information, the role of each entity pair from a global perspective could also be reserved to complement the local structure information.

\subsection{Counterfactual Relations}
\label{sec:CFTriplets}
Once we have defined the treatment from the observed space, we can discover the counterfactual treatment $\gT^{CF}$ and outcome $\gA^{CF}$ for each pair of entities given a query relation. However, this counterfactual relationship is not available from the observed data.
As a substitution, we would like to find $\gT^{CF}$ and $\gA^{CF}$ from the nearest observed context as substitution ~\cite{zhao2022learning}. 
It is a common practice to estimate treatment effects from observational data using this form of matching on covariates~\citep{johansson2016learning,alaa2019validating,zhao2022learning}.
In this way, for a pair of entities $(h, t)$ given a relation $r$ as a candidate, we can match their counterfactual treatment $\gT^{CF}_{(h, r, t)}$ to their nearest neighbor with the opposite treatment. Accordingly, the outcome of their nearest neighbor is treated as \emph{counterfactual relation}.

Formally, $\forall (h_i, r_j, t_k) \in \gS$, where $\gS$ is a selected set of entity pair candidates, its counterfactual relation $(h_a, r_j, t_b)$ with the same type of relationship is
\begin{align}
\label{eq:nearest1}
    (h_a, r_j, t_b) = \argmin_{(h_a, r_j, t_b) \in \gE \times \gE} \big\{&d((h_i, r_j, t_k), (h_a, r_j, t_b)) \big| \notag \\
    & \text{ } \gT^{F}_{(h_a, r_j, t_b)} = 1- \gT^{F}_{(h_i, r_j, t_k)} \big\},
\end{align}
where $d(\cdot, \cdot)$ is a distance measuring metric between a pair of entity pairs (a pair of contexts) in the factual adjacency tensor $\gA^{F}_{r_j}$ given a query relation $r_j$. Here we omit the index in relation $r$ since the counterfactual relation should have the same relation type as the factual one.
Nevertheless, there are still two challenges remaining for the above practice: 
\begin{itemize}[leftmargin=*]
    \item It needs $O(N^4)$ comparisons for finding the nearest neighbors if we choose the commonly used all $O(N^2)$ entity pairs as $\gS = \gE \times \gE$, which is inefficient and infeasible in the application and hard to scale to large graphs. Considering the validity does not hold for all entity pairs, we design to set $\gS$ as all entity pairs that have appeared in the dataset, but we eliminate the types of relations as well as the validity of the formed triplets, which prevents the risk of data leakage. Thus, the number of comparisons is greatly reduced by this selection of $\gS$.
    
    \item Directly calculating the closest distance of entity pairs from $\gA^{F}$ is still trapped in the observed space and could not bring extra counterfactual information. Therefore, we propose to find the nearest observed context of entity pairs from hidden low-dimensional embedding space. We take the unsupervised embedding method node2vec~\cite{grover2016node2vec} to learn the embedding space for entities since it could effectively preserve the proximity in the graph. This aligns with our second reason for choosing K-core for defining factual treatment in the observed space. 
\end{itemize}

After the aforementioned two challenges are solved, we propose to use an entity embedding matrix, which is denoted as $\mM$, to find the counterfactual substitutes of triplets from candidates. 
Especially, taking the imbalanced distribution into account, we design to learn $\mM$ w.r.t the proportion $\psi_j$ of a given relation $r_j$ in the distribution of all relation types. That is for every $\gA^{F}_{r_j}$, we employ node2vec to get the embeddings accordingly as $\mM_{r_j}$, and the overall embedding matrix is defined as:
\begin{equation}
    \mM \in \mathbb{R}^{\gE \times d} = \sum_{r_j}^{\gR} \psi_j \mM_{r_j}. \notag
\end{equation}
where $d$ is the dimension of embedding.
Thus, $\forall (h_i, r_j, t_k) \in \gS$, we define its substitution for counterfactual relation $(h_a, r_j, t_b)$ as
\begin{align}
    \label{eq:nearest2}
    (h_a, r_j, t_b) = \argmin_{(h_a, r_j, t_b) \in \gE \times \gE} \big\{ & d(\vm_{i}, \vm_{a}) + d(\vm_{k}, \vm_{b}) \big| \notag \\
    & \text{ } \gT^{F}_{(h_a, r_j, t_b)} = 1- \gT^{F}_{(h_i, r_j, t_k)} \big\},
\end{align}
where we specify $d(\cdot, \cdot)$ as the Euclidean distance on the hidden embedding space of $\mM$, and $\vm_{i}$ indicates the embedding of entity $h_i$ in the weighted node2vec space and so on. It is worth mentioning that in comparison with the approximated solution in ~\cite{zhao2022learning}, our introduced pair candidates could not only enhance the validity of potential triplets but also cast off the need for another hyperparameter for constraining the maximum distance.

To guarantee that all neighbors are sufficiently comparable (as substitutes) in the hidden embedding space, we do not assign counterfactual treatments for the given entity pair if there does not exist any entity pair satisfying Eq.(\ref{eq:nearest2}), Thus, $\forall (h_i, r_j, t_k) \in \gS$ the counterfactual treatment matrix $\gT^{CF}$ and the counterfactual adjacency tensor $\gA^{CF}_{r_j}$ given a query relation $r_j$ are substituted as
\begin{equation}
\label{eq:tcf}
    \gT^{CF}_{(h_i, r_j, t_k)}, \gA^{CF}_{i, j, k} = 
    \begin{cases}
        \gT^{F}_{(h_a, r_j, t_b)}, \gA^{F}_{a, j, b} & \text{, if } \exists$ $(h_a, r_j, t_b) \in \gE \times \gE \\
         & \quad \text{ satisfies \ref{eq:nearest2};} \\
        \gT^{F}_{(h_i, r_j, t_k)}, \gA^{F}_{i, j, k} &  \text{, otherwise}. 
    \end{cases}
\end{equation}
Note that the embeddings $\mM$ from hidden space 
is only used once for finding the substitutes for counterfactual treatment and outcome. Both $\mM$ and the nearest neighbors are also computed only once and remain unchanged during the learning process.

After the instantiation of the causal model on KGC, we can design our model on how to learn the embeddings for the KG from the counterfactual distributions, which we denote as \method. We denote $P^{F}$ as the {factual distribution} of the observed contexts and treatments, and $P^{CF}$ be the {counterfactual distribution} that is composed of the observed contexts and {opposite} treatments. For all possible triplets $(h_i, r_j, t_k)$ and their treatments, the {empirical} factual distribution $\hat{P}^{F} \sim P^{F}$ is defined as $\hat{P}^{F} = \{((h_i, r_j, t_k), \gT^{F}_{(h_i, r_j, t_k)})\}^\gE_{i,k=1}$, and define the empirical counterfactual distribution $\hat{P}^{CF} \sim P^{CF}$ as $\hat{P}^{CF} = \{((h_i, r_j, t_k), \gT^{CF}_{(h_i, r_j, t_k)})\}^\gE_{i,k=1}$. 
While existing GNN-based KGC models only use the information from $\hat{P}^{F}$ as input, 
we benifit from the counterfactual distribution $\hat{P}^{CF}$ by augment the training data with the counterfactual relations.

\begin{table*}[!h]
    \centering
    \caption{KGC results. Results of NeuraLP and DRUM are taken from \cite{sadeghian2019drum}. Results of RotatE, HAKE, and LowFER are taken from their original papers~\cite{sun2019rotate, zhang2020learning, amin2020lowfer}. Results of all GNNs-based models are reproduced with DGL~\cite{dgl}. Results of the other embedding methods are taken from \cite{sun2019rotate,zhu2021neural}.}
    \label{tab:kg_result}
    \resizebox{0.9\textwidth}{!}{
        \begin{tabular}{llcccccccccc}
            \toprule
            \multirow{2}{*}{\bf{Class}} & \multirow{2}{*}{\bf{Method}}
            & \multicolumn{5}{c}{\bf{FB15k-237}} & \multicolumn{5}{c}{\bf{WN18RR}} \\
            & & \bf{MRR} & \bf{MR} & \bf{H@1} & \bf{H@3} & \bf{H@10} & \bf{MRR} & \bf{MR} & \bf{H@1} & \bf{H@3} & \bf{H@10} \\
            \midrule
            \multirow{3}{*}{\bf{Path-based}}
            & Path Ranking~\cite{lao2010relational} & 0.174 & 3521 & 0.119 & 0.186 & 0.285 & 0.324 & 22438 & 0.276 & 0.360 & 0.406 \\
            & NeuralLP~\cite{yang2017differentiable} & 0.240 & - & - & - & 0.362 & 0.435 & - & 0.371 & 0.434 & 0.566 \\
            & DRUM~\cite{sadeghian2019drum} & 0.343 & - & 0.255 & 0.378 & 0.516 & 0.486 & - & 0.425 & 0.513 & 0.586 \\
            \midrule
            \multirow{8}{*}{\bf{Embeddings}}
            & TransE~\cite{bordes2013translating} & 0.294 & 357 & - & - & 0.465 & 0.226 & 3384 & - & - & 0.501 \\
            & DistMult~\cite{yang2015embedding} & 0.241 & 254 & 0.155 & 0.263 & 0.419 & 0.429 & 5110 & 0.391 & 0.442 & 0.490 \\
            & ComplEx~\cite{trouillon2016complex} & 0.247 & 339 & 0.158 & 0.275 & 0.428 & 0.438 & 5261 & 0.412 & 0.464 & 0.512 \\
            & RotatE~\cite{sun2019rotate} & 0.338 & 177 & 0.241 & 0.375 & 0.553 & 0.476 & 3340 & 0.428 & 0.492 & 0.571 \\
            & HAKE~\cite{zhang2020learning} & 0.346 & - & 0.250 & 0.381 & 0.542 & 0.497 & - & 0.452 & 0.516 & 0.582 \\
            & LowFER~\cite{amin2020lowfer} & 0.359 & - & 0.266 & 0.396 & 0.544 & 0.465 & - & 0.434 & 0.479 & 0.526 \\
            & TuckER~\cite{balavzevic2019tucker}    & 0.358 & - & 0.266 & 0.394 & 0.544 & 0.470 & - & 0.433 & 0.482 & 0.526\\
            & ConvE~\cite{dettmers2018convolutional} & 0.319 & 276 & 0.232 & 0.351 & 0.492 & 0.462 &4888 & 0.431 & 0.476 & 0.525 \\
            \midrule
            \multirow{13}{*}{\bf{GNNs}}
            & RGCN~\cite{schlichtkrull2018modeling} + TransE  & 0.324 & 220 & 0.238 & 0.357 & 0.497 & 0.198 & 3272 & 0.066 & 0.263 & 0.478 \\
            & RGCN~\cite{schlichtkrull2018modeling} + DistMult & 0.332 & 242 & 0.242 & 0.366 &  0.508 & 0.393 & 6980 & 0.350 & 0.413 & 0.481 \\
            & RGCN~\cite{schlichtkrull2018modeling} + ConvE &  0.336 & 197 & 0.249 & 0.368 & 0.511 & 0.428 & 2875 & 0.370 & 0.460 & 0.527 \\
            \cmidrule{2-12}
            & CompGCN~\cite{vashishth2019composition} + TransE  & 0.335 &205 & 0.247 & 0.369 & 0.511 & 0.206 &3182 & 0.064 & 0.281 & 0.502 \\
            & CompGCN~\cite{vashishth2019composition} + DistMult & 0.342 &200 & 0.252 & 0.372 &  0.520 & 0.430 & 4559 & 0.395 & 0.439 & 0.513 \\
            & CompGCN~\cite{vashishth2019composition} + ConvE &  0.351 & 245 & 0.254 & 0.386 & 0.535 & 0.469 & 3065 & 0.433 & 0.480 & 0.543 \\
            \cmidrule{2-12}
            & LTE~\cite{zhang2022rethinking} + TransE  & 0.334 &182 & 0.241 & 0.370 & 0.519 & 0.211 &3290 & 0.022 & 0.362 & 0.521\\
            & LTE~\cite{zhang2022rethinking} + DistMult & 0.335 &238 & 0.246 & 0.367 &  0.517  & 0.437 & 4485 & 0.403 & 0.447 & 0.517 \\
            & LTE~\cite{zhang2022rethinking} + ConvE & 0.352 & 249 & 0.262 & 0.385 & 0.533  & 0.472 &3434 & 0.436 & 0.485 & 0.544\\
            \cmidrule{2-12}
            & NBFNet~\cite{zhu2021neural} & 0.415 & \textbf{114} & 0.321 & 0.454 & 0.599 & 0.551 & 636 & 0.497 & 0.573 & 0.666 \\
            & \method (\textbf{Ours}) & \best{0.421} & 124 & \best{0.326} & \best{0.461} & \best{0.603} & \best{0.553} & \best{628} & \best{0.503} & \best{0.577} & \best{0.670} \\
            \bottomrule
        \end{tabular}
    }
\end{table*}

\subsection{Counterfactual Relations Augmented Learning Framework}
\label{sec:model}
After all the data preparation, as with most of the GNN-based embedding model, our model \method consists of two trainable components: a KG encoder $f$ and a relation decoder $g$.

\xhdr{Encoder}
While we aim to learn a pairwise representation $\vz_r(h, t)$, the encoder $f$ should capture the important neighborhood information between $h$ and $t$ w.r.t the query relation $r$, such as the proximity that is preserved by counting different types of random walks similar to the one we use during finding counterfactual relations. This kind of encoded local structure is often captured by path formulation, which is widely used in KG representation learning~\cite{lao2010relational,gardner2015efficient,li2019predicting,zhu2021neural}
Without the loss of generality, we adopt the most advanced variant NBFNet~\cite{zhu2021neural} as the encoder $f$ in \method.

NBFNet is a path formulation-based graph representation learning framework. In NBFNet, the pair representations of nodes are interpreted as a generalized sum and each path representation is defined as the generalized product of the edge representations in the path. Then NBFNet solves the path formulation efficiently with learned operators with the help of the generalized Bellman-Ford algorithm,
then reads out the pair representation as a source-specific message-passing process.
In NBFNet, each layer of GNN is defined as
\begin{align}\label{eq:gcn_layer}
    \vspace{-6mm}
    \vz^{(0)}_r(h,t) &\leftarrow \textsc{IND}(h, r, t), \notag \\
    \vz^{(l+1)}_r(h,t) &\leftarrow \textsc{AGG}\Big(\Big\{\textsc{MSG}\big(\vz^{(l)}_r(h,t), \vw(h, r, e)\big) \big| \notag \\
    & \qquad \qquad \qquad (h, r, e) \in \mathcal{N}_{in}(e) \Big\} \cup \Big\{\vz^{(0)}_r(h,t)\Big\}\Big)
    \vspace{-6mm}
\end{align}
where $l$ is the layer index. $\vz^{(l)}_r(h,t)$ is the pair representation from $l$ layer for entity pair $(h,t)$ given a query relation $r$. The set $\mathcal{N}_{in}(e)$ contains all triplets $(h,r,e)$ such that $(h,r,e)$ is valid for a preset entity $e$. $\vw(h, r, e)$ is the edge representation for the triplet $(h, r, e)$. 
The \textsc{IND} is the indicator function outputs $\oone_r$ if $h = t$ and $\ozero_r$ otherwise, which aims to provide a non-trivial representation for the heat entity $h$ as the boundary condition.
The \textsc{AGG} function is instantiated as natural sum, max or min in traditional methods followed by a linear transformation and a non-linear activation, or principal neighborhood aggregation (PNA) in recent efforts~\cite{corso2020principal}. For the \textsc{MSG} function, the NBFNet incorporates the traditional score functions (\eg, TransE, DistMult, or RotatE) as transformations correspond to the relational operators, in order to model the message passing procedure among entities and relations.

\xhdr{Decoder}
For the relation decoder that predicts whether a triplet is valid between a pair of entities with a query relation $r$, we remain consistent with ~\cite{zhao2022learning} and adopt a simple decoder based on multi-layer perceptron (MLP) to predict the conditional likelihood of the tail entity $t$. With the pair representation $\vz_r(h,t)$ from the output of the last layer in encoder $f$ and the treatments of the triplet, the decoder $g$ is defined as
\begin{align}
    \label{eq:apredf}
    &\hat{p}^{F}(t | h, r) = g(\gZ, \gT^{F}) = \text{MLP}([\vz_r(h,t), \gT^{F}_{(h, r, t)}]), \\
    \label{eq:apredcf}
    &\hat{p}^{CF}(t | h, r) = g(\gZ, \gT^{CF}) = \text{MLP}([\vz_r(h,t), \gT^{CF}_{(h, r, t)}]),
\end{align}
where $[\cdot, \cdot]$ stands for the concatenation of vectors.
The outputs are the conditional likelihood of the tail entity $t$: $\hat{p}^{F}(t | h, r)$ and $\hat{p}^{CF}(t | h, r)$ for both factual and counterfactual empirical distribution. The conditional likelihood of the head entity $h$ can be predicted by $\hat{p}^{F}(h|t, r^{-1})$ and $\hat{p}^{CF}(h|t, r^{-1})$ with the same model. Note that only the factual output $\hat{p}^{F}(t | h, r)$ will be used in inference.

\xhdr{Loss design}
During the training process, data samples from the empirical factual/counterfactual distribution $\hat{P}^{F}$/$\hat{P}^{CF}$ are fed into decoder $g$ and optimized towards $p^{F}(t | h, r)$ and $p^{CF}(t | h, r)$, respectively. We follow the previous works~\cite{bordes2013translating, sun2019rotate} to minimize the negative log-likelihood of positive and negative triplets from factual and counterfactual views. The negative samples are generated according to Partial Completeness Assumption (PCA)~\cite{galarraga2013amie} as in NBFNet, which creates a negative sample by corrupting one of the entities in a positive triplet. The corresponding losses are defined as:
\begin{align}
    \label{eq:eplossf}
    \mathcal{L}_{F} = &-\log \hat{p}^{F}(h, r, t) - \sum_{i=1}^n \frac{1}{n}\log (1 - \hat{p}^{F}(h_i', r, t_i')), \\
    \label{eq:eplosscf}
    \mathcal{L}_{CF} = &-\log \hat{p}^{CF}(h, r, t) - \sum_{i=1}^n \frac{1}{n}\log (1 - \hat{p}^{CF}(h_i', r, t_i')).
\end{align}
where $n$ is the number of negative samples for a single positive sample and $(h_i', r, t_i')$ is the $i$-th negative sample. The negative samples remain the same for factual and counterfactual treatments.

Since the test data contains only observed (factual) samples, we force the distributions of representations of factual distributions and counterfactual distributions to be similar, which could protect the model from exposing to the risk of covariant shift~\cite{johansson2016learning,assaad2021counterfactual}. Similar to ~\cite{zhao2022learning}, we minimize the discrepancy distance~\cite{mansour2009domain} between $\hat{P}^{F}$ and $\hat{P}^{CF}$ to regularize the representation learning:
\begin{equation}
    \label{eq:discloss}
    \mathcal{L}_{disc} = \text{disc}(\hat{P}^{F}_f, \hat{P}^{CF}_f), \text{ where } \text{disc}(P, Q) = ||P - Q||_{F},
\end{equation}
where $||\cdot||_F$ denotes the Frobenius Norm. $\hat{P}^{F}_f$ and $\hat{P}^{CF}_f$ are the representations of entity pairs given a query relation learned by graph encoder $f$ from factual distribution and counterfactual distribution, respectively.
Therefore, the overall training loss of our proposed \method is
\begin{equation}
    \mathcal{L} = \mathcal{L}_{F} + \alpha \cdot \mathcal{L}_{CF} + \beta \cdot \mathcal{L}_{disc},
    \label{eq:loss}
\end{equation}
where $\alpha$ and $\beta$ are hyperparameters to control the weights of counterfactual outcome estimation loss and discrepancy loss.
Recall from Sec.~\ref{sec:causalmodel} that our estimated $\widehat{\text{ATE}}$ is formulated as:
\begin{align}
    \widehat{\text{ATE}} = &\frac{1}{\gE \times \gE} \sum_{i=1}^{\gE} \sum_{j=1}^{\gE} {\{} \gT^{F} \odot (\gA^{F} - \gA^{CF}) \\
    &+ (\mathbf{1}_{\gE \times \gE} - \gT^{F}) \odot (\gA^{CF} - \gA^{F}) {\}}_{i,j}.
\end{align} 
We can find from loss Eq.(\ref{eq:loss}) that $\widehat{\text{ATE}}$ is implicitly optimized during learning from the counterfactual relations. As a result, the learned causal relations help to improve the learning of pair representation.

\xhdr{Computational complexity}
We first analyze the computational complexity of finding counterfactual relations. The complexity of the node2vec for hidden embedding space is bounded by $\mathcal{O} (\gE\gR)$. The step of finding counterfactual links with nearest neighbors in Eq.(\ref{eq:nearest2}) has complexity in proportion to the size of the candidate set $|S|$ and entity pairs. Since the computation in \ref{eq:nearest2} can be parallelized, the time complexity is $O(|S|\gE^2/C)$ where $C$ is the number of processes.
Then for the complexity of the training counterfactual learning model, the NBFNet encoder $f$ has a time complexity of $\mathcal{O} (\gE D_f + \gH D_f^2 )$, where $\gH$ is the number of training triplets and $D_f$ is the dimension of entity representations. While we sample the same number of non-existing triplets as that of observed triplets during training, the complexity of a \emph{two-layer MLP} decoder $g$ is $O(((D_f+1)\cdot D_g + 1 \cdot D_g ) \gH) = O((D_f+2) D_g \gH)$, where $D_g$ is the number of neurons in the MLP hidden layer.

\begin{table*}[!h]
    \centering
    \caption{Examples of path interpretations of predictions on FB15k-237. For a valid triplet as a query, we visualize the \textit{S}ubstitution \textit{o}f its \textit{C}ounterfactual \textit{R}elation, which is denoted as SoCR, as well as the top-1 path interpretations and their weights. Inverse relations are denoted with a superscript $^{-1}$.}
    \label{tab:visualization}
    \vspace{-2mm}
    \resizebox{0.75\textwidth}{!}{
        % \footnotesize
        \begin{tabular}{ll}
            \toprule
            \bf{Query} & \edge{$h_i, r_j, t_k$}: \edge{Barzil, olympics, 1972 summer olympics}, $(\gT^{F}_{h_i, r_j, t_k} = 1$, $\gT^{CF}_{h_i, r_j, t_k} = 0)$ \\
            \bf{SoCR} & \edge{$h_a, r_j, t_b$}: \edge{Argentinia, olympics, Bids for the 1976 summer olympics}  $(\gT^{F}_{h_a, r_j, t_b} = 0$, $\gT^{CF}_{h_a, r_j, t_b} = 0)$ \\
            \midrule
            \multirow{2}{*}{0.350} & \edge{Barzil, medal, Gold Medal} $(\gT^{F}_{h_i, r_j, t_k} = 1$, $\gT^{CF}_{h_i, r_j, t_k} = 0)$ \\
            & $\land$ \edge{Gold Medal, medal$^{-1}$, 1972 summer olympics} $(\gT^{F}_{h_i, r_j, t_k} = 1$, $\gT^{CF}_{h_i, r_j, t_k} = 0)$ \\
            \midrule
            \bf{Query} & \edge{$h_i, r_j, t_k$}: \edge{Bridesmaids, film release region, Republica Portuguesa}, $(\gT^{F}_{h_i, r_j, t_k} = 0$, $\gT^{CF}_{h_i, r_j, t_k} = 1)$ \\
            \bf{SoCR} & \edge{$h_a, r_j, t_b$}: \edge{The Hangover, film release region, Norwegen}  $(\gT^{F}_{h_a, r_j, t_b} = 1$, $\gT^{CF}_{h_a, r_j, t_b} = 0)$\\
            \midrule
            \multirow{2}{*}{0.445} & \edge{Bridesmaids, film release region, Espagna} $(\gT^{F}_{h_i, r_j, t_k} = 0$, $\gT^{CF}_{h_i, r_j, t_k} = 1)$ \\
            & $\land$ \edge{Espagna, Adjoins$^{-1}$, Republica Portuguesa} $(\gT^{F}_{h_i, r_j, t_k} = 1$, $\gT^{CF}_{h_i, r_j, t_k} = 0)$ \\

            \bottomrule
        \end{tabular}
    }
    \vspace{-2mm}
\end{table*}

\section{Experiments}\label{sec:exps}
\subsection{Experimental Setup}
We evaluate \method on the task KGC on the most popular datasets: FB15k-237~\cite{toutanova2015observed} and WN18RR~\cite{dettmers2018convolutional}.
Both datasets are publicly accessible from DGL\footnote{https://github.com/awslabs/dgl-ke/tree/master/examples}~\cite{dgl}.
We use the standard splits of both datasets.
Statistics of datasets can be found in the Appendix.

\xhdr{Baselines} We compare \method against 21 baselines including three types: path-based methods, embedding methods, and GNNs with different scoring functions. Note that LTE~\cite{zhang2022rethinking} is not exactly a GNN-based model but we include it here since it shares the same design framework with other GNN-based models. 

\xhdr{Implementation Details.} Our implementation is similar with~\cite{zhu2021neural}. 
For the encoder, we closely follow the official implementation for NBFNet and keep all architecture details as reported to maintain a fair comparison. In NBFNet, we set GNN layers as $6$ and hidden states as $32$. The negative ratio (\#negative/\#positive) is set to $32$. For other details please kindly refer to~\cite{zhu2021neural}. For the decoder, we set MLP as two layers with $64$ hidden units and use ReLU as the activation function. The integer $k$ in K-core is set to $2$. The $\text{disc}(\cdot)$ function is selected as KL divergence.
We use a public implementation for node2vec and the output dimension for node2vec is set to $32$. We 
decide the hyperparameters in loss through grid search within: $\alpha \in \{0.001, 0.01, 0.1, 1\}$ and $\beta \in \{0.001, 0.01, 0.1, 1\}$. The batch size is set to $32$. We use the Adam optimizer with a learning rate $5e-3$ for optimization.
Our model is trained on a single Tesla V100 GPU for 20 epochs and reports the results from the model selected based on the best performance on the validation set. 

\subsection{KGC Results}
Table~\ref{tab:kg_result} shows the performance of \method on the task of KGC in comparison with state-of-the-art baselines. We report the results regarding the filtered ranking protocol~\cite{bordes2013translating}. We have the following observations: \textbf{1)} Our \method achieves the best performance across all measurements on both datasets, which aligns with our motivation that balancing the distribution of relation types with augmentation could contribute to better pair representation learning. \textbf{2)} The path-based GNN approaches generally outperform other GNN baselines, which implies that the local subgraph structure that maintains the proximity between an entity pair is more essential to the massage-passing mechanism in the GNN-based KGC framework.
This is consistent with the observation from LTE~\cite{zhang2022rethinking} that the contribution of current convolutional GNN-based designs to KGC is somehow unnecessary and needs further exploration. \textbf{3)} We further use a two-sided \textit{t}-test to evaluate the significance of KGCF over NBFNet. Due to limited space here, we only show the result of the p-value on WN18RR here:
\begin{table}[htbp]
\renewcommand{\arraystretch}{1.11}
\caption{Significance test of KGCF over NBFNet on WN18RR.} \label{tab:ttest}
\vspace{-3mm}
\begin{center}
\resizebox{0.75\columnwidth}{!}{
\begin{tabular}{ c c c c c c c}
\hline
Metric  & \bf{MRR} & \bf{H@1} & \bf{H@3} & \bf{H@10} \\
\hline
p-value   & P<0.005 & P<0.001 & P<0.001 & P<0.005 \\
\hline
\end{tabular}
}
\end{center}
\vspace{-4mm}
\end{table}
We can find that KGCF achieves strongly significant improvements (P < 0.005) over NBFNet across all metrics, which statistically confirms that KGCF's scores in Table~\ref{tab:kg_result} are significantly better than NBFNet's.

\subsection{Ablation Studies}
\xhdr{Visualization of prediction}
With the help of the interpretation tool from NBFNet, we could interpret the KGC prediction $\hat{p}^{F}(h, r, t)$ of \method through paths from $h$ to $t$ that contribute most to the results. 
The importance of a path interpretation for $\hat{p}^{F}(h, r, t)$ is defined as its weight from a linear model, where the weight is approximated by the sum of the importance of edges in that path. In other words, the $\topk$ path interpretations are equivalent to the $\topk$ longest paths on the edge importance graph. The is computed by the partial derivative of the prediction w.r.t the path:
\begin{equation}
    \gP_1, \gP_2, ..., \gP_k = \topk_{\gP \in \gP_{h,t}} \frac{\partial{\hat{p}^{F}(h, r, t)}}{\partial{\hat{P}^{F}}},
\end{equation}
where $\gP_1, \gP_2, ..., \gP_k$ is the $\topk$ paths acquired from ranking weights. 

Table~\ref{tab:visualization} shows the interpretation of two triplets examples with the path as well as their substitutions that are used for calculating counterfactual treatment SoCR. 
Note that SoCR does not have to be a valid triplet. Either way, its validity could augment the learning with a counterfactual view of information. This augmentation helps avoid inaccurate predictions through directly analogical reasoning, which is the common practice of previous efforts, from only observed factual neighborhoods.

Our observations from Table~\ref{tab:visualization} are: \textbf{1)} The reasoning over two entities is conducted over the paths among them with both information from factual and counterfactual treatments. Our method provides two additional views of the learned factual and counterfactual relations, which helps enhance the confidence level of reasoning by answering the counterfactual question. \textbf{2)} The nearest observed context that is used as a substitution for counterfactual estimation is effective on the KG. We can find that similar entities (\eg, the films Bridesmaids and The Hangover) are close to each other in the proximity-preserving embedding space and with analogous context. \textbf{3)} The interpretation answers our question from Section~\ref{sec:intro} that the relationship could not exist when the neighborhood information varies. In the first example, Argentina did not bid for the 1976 summer Olympics thus the relation is not valid in counterfactual treatment. This augmented counterfactual view of information helps the learning avoid false prediction because analogical reasoning from only observed factual neighborhoods might infer it as a valid one otherwise.

\xhdr{Running time comparison}
In complementary of the complexity analysis from Section~\ref{sec:model}, we report the wall time on the FB15k-237 dataset for inference efficiency comparison over a single Tesla V100 GPU. It is obvious that the \method is very comparable to the NBFNet encoder, which indicates that our augmented framework brings no burden to the existing GNN-based method. Meanwhile, we can find that the convolutional GNN-based methods are the most efficient since they do not consider a path-based strategy, which results in worse KGC performance conversely. 

\begin{table}[htbp]
\renewcommand{\arraystretch}{1.11}
\caption{Inference time comparison on FB15k-237 test set with a single GPU. We report the average time over $5$ splits for each method here. $s$ denotes seconds and $m$ denotes minutes.} \label{tab:timeCost}
\vspace{-2mm}
\begin{center}
\resizebox{\columnwidth}{!}{%
\begin{tabular}{ c c c c c c c}
\hline
Methods  & RGCN & CompGCN & LTE & NeuralLP & NBFNet & \method \\
\hline
Wall Time    & 13$s$ & 5$s$ & 11$s$ & 2.1$m$  & 4.0$m$ & 4.2$m$   \\
\hline
\end{tabular}
}
\end{center}
\end{table}

\begin{figure}[tb]
\centering
\vspace{-2mm}
\includegraphics [width=0.95\columnwidth]{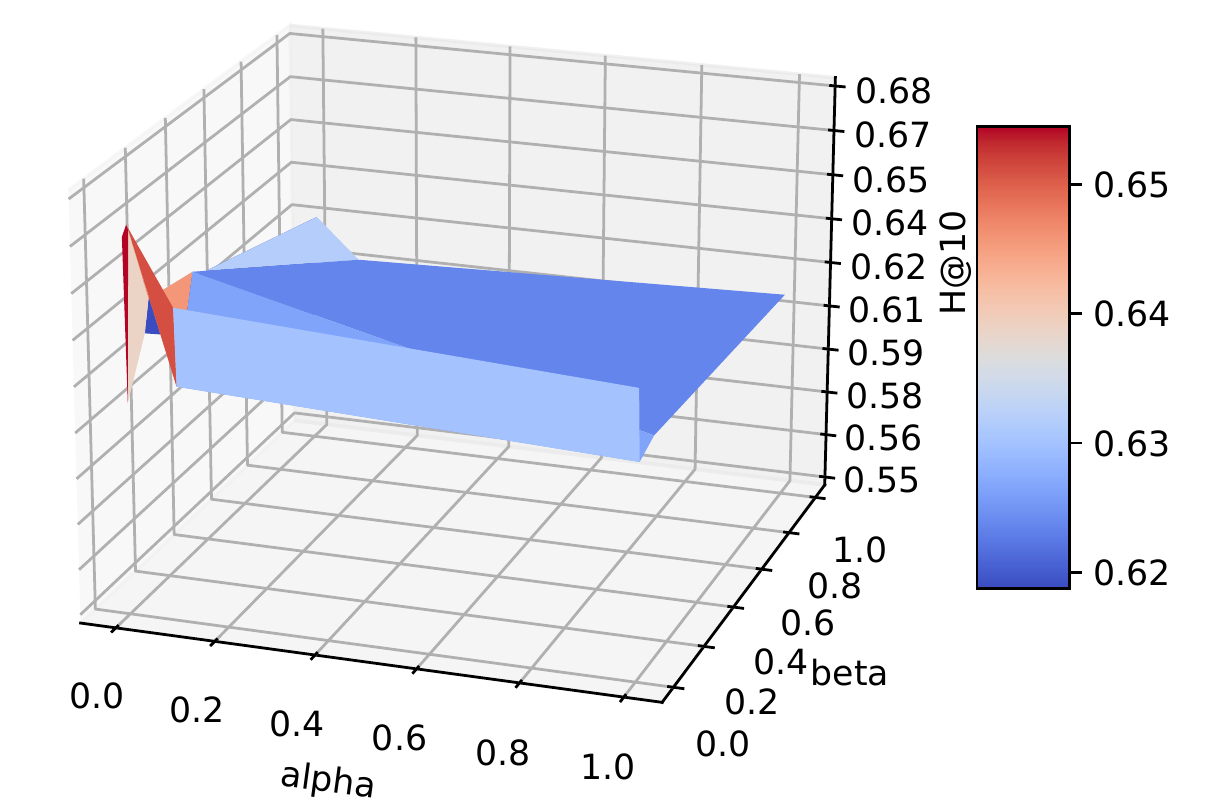}
\vspace{-3mm}
\caption{Sensitivity analysis on WN18RR w.r.t $\alpha$ and $\beta$.}
\Description{Sensitivity analysis on WN18RR w.r.t $\alpha$ and $\beta$. We observe that the performance is relatively stable and \method is generally robust to the hyperparameters $\alpha$ and $\beta$ for finding the optimal values.}
\vspace{-6mm}
\label{fig:sensitivity}
\end{figure}

\xhdr{Ablation study on $\mathcal{L}_{CF}$ and $\mathcal{L}_{disc}$}
To investigate how $\mathcal{L}_{CF}$ and $\mathcal{L}_{disc}$ affect the overall KGC performance, we conduct sensitivity analysis on the performance of \method w.r.t different combinations of $\alpha$ and $\beta$. While we tune $\alpha$ and $\beta$ through grid search within the same set $\{0.001, 0.01, 0.1, 1\}$, we search through all combinations on WN18RR and report the corresponding H@10 performance in Figure~\ref{fig:sensitivity}. We observe that the performance is relatively stable and \method is generally robust to the hyperparameters $\alpha$ and $\beta$ for finding the optimal values.

\section{Conclusion}\label{sec:conclusion}
In this paper, we propose a novel counterfactual relation-based data augmentation framework \method for the task of KGC. Given a pair of entities and a query relation, we first explore the factual treatments for the observed context from the view of the proximity of entities. Considering the imbalance of the distribution of relation types, we define the counterfactual relations with the help of their nearest neighbors in the embedding space from relation type weighted node2vec. The counterfactual relations augment the training data with causal relationships and facilitate pair representation learning for KGC. Comprehensive experiments demonstrate that \method achieves the SOTA performance on benchmarks as well as endows the interpretability of the path-based GNN models on KGC from a new causal perspective.

\section*{Acknowledgments}\label{sec:ack}
% \begin{ack}
This work was partially supported by NSFC Grant No. 62206067 and HKUST-GZU Joint Research Collaboration Fund No.GZU22EG05.
% \end{ack}

\bibliographystyle{ACM-Reference-Format}
\bibliography{refs}

%%% -*-BibTeX-*-
%%% Do NOT edit. File created by BibTeX with style
%%% ACM-Reference-Format-Journals [18-Jan-2012].

\begin{thebibliography}{58}

%%% ====================================================================
%%% NOTE TO THE USER: you can override these defaults by providing
%%% customized versions of any of these macros before the \bibliography
%%% command.  Each of them MUST provide its own final punctuation,
%%% except for \shownote{}, \showDOI{}, and \showURL{}.  The latter two
%%% do not use final punctuation, in order to avoid confusing it with
%%% the Web address.
%%%
%%% To suppress output of a particular field, define its macro to expand
%%% to an empty string, or better, \unskip, like this:
%%%
%%% \newcommand{\showDOI}[1]{\unskip}   % LaTeX syntax
%%%
%%% \def \showDOI #1{\unskip}           % plain TeX syntax
%%%
%%% ====================================================================

\ifx \showCODEN    \undefined \def \showCODEN     #1{\unskip}     \fi
\ifx \showDOI      \undefined \def \showDOI       #1{#1}\fi
\ifx \showISBNx    \undefined \def \showISBNx     #1{\unskip}     \fi
\ifx \showISBNxiii \undefined \def \showISBNxiii  #1{\unskip}     \fi
\ifx \showISSN     \undefined \def \showISSN      #1{\unskip}     \fi
\ifx \showLCCN     \undefined \def \showLCCN      #1{\unskip}     \fi
\ifx \shownote     \undefined \def \shownote      #1{#1}          \fi
\ifx \showarticletitle \undefined \def \showarticletitle #1{#1}   \fi
\ifx \showURL      \undefined \def \showURL       {\relax}        \fi
% The following commands are used for tagged output and should be
% invisible to TeX
\providecommand\bibfield[2]{#2}
\providecommand\bibinfo[2]{#2}
\providecommand\natexlab[1]{#1}
\providecommand\showeprint[2][]{arXiv:#2}

\bibitem[\protect\citeauthoryear{Alaa and Van Der~Schaar}{Alaa and Van
  Der~Schaar}{2019}]%
        {alaa2019validating}
\bibfield{author}{\bibinfo{person}{Ahmed Alaa} {and} \bibinfo{person}{Mihaela
  Van Der~Schaar}.} \bibinfo{year}{2019}\natexlab{}.
\newblock \showarticletitle{Validating causal inference models via influence
  functions}. In \bibinfo{booktitle}{\emph{International Conference on Machine
  Learning}}. \bibinfo{pages}{191--201}.
\newblock


\bibitem[\protect\citeauthoryear{Amin, Varanasi, Dunfield, and Neumann}{Amin
  et~al\mbox{.}}{2020}]%
        {amin2020lowfer}
\bibfield{author}{\bibinfo{person}{Saadullah Amin}, \bibinfo{person}{Stalin
  Varanasi}, \bibinfo{person}{Katherine~Ann Dunfield}, {and}
  \bibinfo{person}{G{\"u}nter Neumann}.} \bibinfo{year}{2020}\natexlab{}.
\newblock \showarticletitle{LowFER: Low-rank bilinear pooling for link
  prediction}. In \bibinfo{booktitle}{\emph{International Conference on Machine
  Learning}}. PMLR, \bibinfo{pages}{257--268}.
\newblock


\bibitem[\protect\citeauthoryear{Assaad, Zeng, Tao, Datta, Mehta, Henao, Li,
  and Duke}{Assaad et~al\mbox{.}}{2021}]%
        {assaad2021counterfactual}
\bibfield{author}{\bibinfo{person}{Serge Assaad}, \bibinfo{person}{Shuxi Zeng},
  \bibinfo{person}{Chenyang Tao}, \bibinfo{person}{Shounak Datta},
  \bibinfo{person}{Nikhil Mehta}, \bibinfo{person}{Ricardo Henao},
  \bibinfo{person}{Fan Li}, {and} \bibinfo{person}{Lawrence~Carin Duke}.}
  \bibinfo{year}{2021}\natexlab{}.
\newblock \showarticletitle{Counterfactual representation learning with
  balancing weights}. In \bibinfo{booktitle}{\emph{International Conference on
  Artificial Intelligence and Statistics}}. PMLR, \bibinfo{pages}{1972--1980}.
\newblock


\bibitem[\protect\citeauthoryear{Bala{\v{z}}evi{\'c}, Allen, and
  Hospedales}{Bala{\v{z}}evi{\'c} et~al\mbox{.}}{2019}]%
        {balavzevic2019tucker}
\bibfield{author}{\bibinfo{person}{Ivana Bala{\v{z}}evi{\'c}},
  \bibinfo{person}{Carl Allen}, {and} \bibinfo{person}{Timothy Hospedales}.}
  \bibinfo{year}{2019}\natexlab{}.
\newblock \showarticletitle{TuckER: Tensor Factorization for Knowledge Graph
  Completion}. In \bibinfo{booktitle}{\emph{Proceedings of the 2019 Conference
  on Empirical Methods in Natural Language Processing and the 9th International
  Joint Conference on Natural Language Processing (EMNLP-IJCNLP)}}.
  \bibinfo{pages}{5185--5194}.
\newblock


\bibitem[\protect\citeauthoryear{Bordes, Usunier, Garcia-Duran, Weston, and
  Yakhnenko}{Bordes et~al\mbox{.}}{2013}]%
        {bordes2013translating}
\bibfield{author}{\bibinfo{person}{Antoine Bordes}, \bibinfo{person}{Nicolas
  Usunier}, \bibinfo{person}{Alberto Garcia-Duran}, \bibinfo{person}{Jason
  Weston}, {and} \bibinfo{person}{Oksana Yakhnenko}.}
  \bibinfo{year}{2013}\natexlab{}.
\newblock \showarticletitle{Translating embeddings for modeling
  multi-relational data}. In \bibinfo{booktitle}{\emph{Advances in Neural
  Information Processing Systems}}. \bibinfo{pages}{1--9}.
\newblock


\bibitem[\protect\citeauthoryear{Cai, Li, Wang, and Ji}{Cai
  et~al\mbox{.}}{2021}]%
        {cai2021line}
\bibfield{author}{\bibinfo{person}{Lei Cai}, \bibinfo{person}{Jundong Li},
  \bibinfo{person}{Jie Wang}, {and} \bibinfo{person}{Shuiwang Ji}.}
  \bibinfo{year}{2021}\natexlab{}.
\newblock \showarticletitle{Line graph neural networks for link prediction}.
\newblock \bibinfo{journal}{\emph{IEEE Transactions on Pattern Analysis and
  Machine Intelligence}} (\bibinfo{year}{2021}).
\newblock


\bibitem[\protect\citeauthoryear{Chang, Rong, Xu, Bian, Zhou, Wang, Huang, and
  Zhu}{Chang et~al\mbox{.}}{2021a}]%
        {chang2021not}
\bibfield{author}{\bibinfo{person}{Heng Chang}, \bibinfo{person}{Yu Rong},
  \bibinfo{person}{Tingyang Xu}, \bibinfo{person}{Yatao Bian},
  \bibinfo{person}{Shiji Zhou}, \bibinfo{person}{Xin Wang},
  \bibinfo{person}{Junzhou Huang}, {and} \bibinfo{person}{Wenwu Zhu}.}
  \bibinfo{year}{2021}\natexlab{a}.
\newblock \showarticletitle{Not All Low-Pass Filters are Robust in Graph
  Convolutional Networks}.
\newblock \bibinfo{journal}{\emph{Advances in Neural Information Processing
  Systems (NeurIPS)}}  \bibinfo{volume}{34} (\bibinfo{year}{2021}).
\newblock


\bibitem[\protect\citeauthoryear{Chang, Rong, Xu, Huang, Sojoudi, Huang, and
  Zhu}{Chang et~al\mbox{.}}{2021b}]%
        {chang2021spectral}
\bibfield{author}{\bibinfo{person}{Heng Chang}, \bibinfo{person}{Yu Rong},
  \bibinfo{person}{Tingyang Xu}, \bibinfo{person}{Wenbing Huang},
  \bibinfo{person}{Somayeh Sojoudi}, \bibinfo{person}{Junzhou Huang}, {and}
  \bibinfo{person}{Wenwu Zhu}.} \bibinfo{year}{2021}\natexlab{b}.
\newblock \showarticletitle{Spectral graph attention network with fast
  eigen-approximation}. In \bibinfo{booktitle}{\emph{Proceedings of the 30th
  ACM International Conference on Information \& Knowledge Management (CIKM)}}.
  \bibinfo{pages}{2905--2909}.
\newblock


\bibitem[\protect\citeauthoryear{Chang, Rong, Xu, Huang, Zhang, Cui, Wang, Zhu,
  and Huang}{Chang et~al\mbox{.}}{2022}]%
        {chang2022adversarial}
\bibfield{author}{\bibinfo{person}{Heng Chang}, \bibinfo{person}{Yu Rong},
  \bibinfo{person}{Tingyang Xu}, \bibinfo{person}{Wenbing Huang},
  \bibinfo{person}{Honglei Zhang}, \bibinfo{person}{Peng Cui},
  \bibinfo{person}{Xin Wang}, \bibinfo{person}{Wenwu Zhu}, {and}
  \bibinfo{person}{Junzhou Huang}.} \bibinfo{year}{2022}\natexlab{}.
\newblock \showarticletitle{Adversarial Attack Framework on Graph Embedding
  Models with Limited Knowledge}.
\newblock \bibinfo{journal}{\emph{IEEE Transactions on Knowledge and Data
  Engineering (TKDE)}} (\bibinfo{year}{2022}).
\newblock


\bibitem[\protect\citeauthoryear{Chang, Rong, Xu, Huang, Zhang, Cui, Zhu, and
  Huang}{Chang et~al\mbox{.}}{2020}]%
        {chang2020restricted}
\bibfield{author}{\bibinfo{person}{Heng Chang}, \bibinfo{person}{Yu Rong},
  \bibinfo{person}{Tingyang Xu}, \bibinfo{person}{Wenbing Huang},
  \bibinfo{person}{Honglei Zhang}, \bibinfo{person}{Peng Cui},
  \bibinfo{person}{Wenwu Zhu}, {and} \bibinfo{person}{Junzhou Huang}.}
  \bibinfo{year}{2020}\natexlab{}.
\newblock \showarticletitle{A restricted black-box adversarial framework
  towards attacking graph embedding models}. In
  \bibinfo{booktitle}{\emph{Proceedings of the AAAI conference on Artificial
  Intelligence (AAAI)}}, Vol.~\bibinfo{volume}{34}.
  \bibinfo{pages}{3389--3396}.
\newblock


\bibitem[\protect\citeauthoryear{Chen, Zhu, Lin, and Cao}{Chen
  et~al\mbox{.}}{2021}]%
        {CHEN2021352}
\bibfield{author}{\bibinfo{person}{Xiangyan Chen}, \bibinfo{person}{Duoduo
  Zhu}, \bibinfo{person}{Dazhen Lin}, {and} \bibinfo{person}{Donglin Cao}.}
  \bibinfo{year}{2021}\natexlab{}.
\newblock \showarticletitle{Rumor knowledge embedding based data augmentation
  for imbalanced rumor detection}.
\newblock \bibinfo{journal}{\emph{Information Sciences}}  \bibinfo{volume}{580}
  (\bibinfo{year}{2021}), \bibinfo{pages}{352--370}.
\newblock
\showISSN{0020-0255}
\urldef\tempurl%
\url{https://doi.org/10.1016/j.ins.2021.08.059}
\showDOI{\tempurl}


\bibitem[\protect\citeauthoryear{Corso, Cavalleri, Beaini, Li{\`o}, and
  Veli{\v{c}}kovi{\'c}}{Corso et~al\mbox{.}}{2020}]%
        {corso2020principal}
\bibfield{author}{\bibinfo{person}{Gabriele Corso}, \bibinfo{person}{Luca
  Cavalleri}, \bibinfo{person}{Dominique Beaini}, \bibinfo{person}{Pietro
  Li{\`o}}, {and} \bibinfo{person}{Petar Veli{\v{c}}kovi{\'c}}.}
  \bibinfo{year}{2020}\natexlab{}.
\newblock \showarticletitle{Principal neighbourhood aggregation for graph
  nets}.
\newblock \bibinfo{journal}{\emph{Advances in Neural Information Processing
  Systems}}  \bibinfo{volume}{33} (\bibinfo{year}{2020}),
  \bibinfo{pages}{13260--13271}.
\newblock


\bibitem[\protect\citeauthoryear{Dettmers, Minervini, Stenetorp, and
  Riedel}{Dettmers et~al\mbox{.}}{2018}]%
        {dettmers2018convolutional}
\bibfield{author}{\bibinfo{person}{Tim Dettmers}, \bibinfo{person}{Pasquale
  Minervini}, \bibinfo{person}{Pontus Stenetorp}, {and}
  \bibinfo{person}{Sebastian Riedel}.} \bibinfo{year}{2018}\natexlab{}.
\newblock \showarticletitle{Convolutional 2d knowledge graph embeddings}. In
  \bibinfo{booktitle}{\emph{Proceedings of the AAAI conference on artificial
  intelligence}}, Vol.~\bibinfo{volume}{32}.
\newblock


\bibitem[\protect\citeauthoryear{Ding, Xu, Tong, and Liu}{Ding
  et~al\mbox{.}}{2022}]%
        {ding2022data}
\bibfield{author}{\bibinfo{person}{Kaize Ding}, \bibinfo{person}{Zhe Xu},
  \bibinfo{person}{Hanghang Tong}, {and} \bibinfo{person}{Huan Liu}.}
  \bibinfo{year}{2022}\natexlab{}.
\newblock \showarticletitle{Data augmentation for deep graph learning: A
  survey}.
\newblock \bibinfo{journal}{\emph{arXiv preprint arXiv:2202.08235}}
  (\bibinfo{year}{2022}).
\newblock


\bibitem[\protect\citeauthoryear{Gal{\'a}rraga, Teflioudi, Hose, and
  Suchanek}{Gal{\'a}rraga et~al\mbox{.}}{2015}]%
        {galarraga2015fast}
\bibfield{author}{\bibinfo{person}{Luis Gal{\'a}rraga},
  \bibinfo{person}{Christina Teflioudi}, \bibinfo{person}{Katja Hose}, {and}
  \bibinfo{person}{Fabian~M Suchanek}.} \bibinfo{year}{2015}\natexlab{}.
\newblock \showarticletitle{Fast rule mining in ontological knowledge bases
  with AMIE ++}.
\newblock \bibinfo{journal}{\emph{The VLDB Journal}} \bibinfo{volume}{24},
  \bibinfo{number}{6} (\bibinfo{year}{2015}), \bibinfo{pages}{707--730}.
\newblock


\bibitem[\protect\citeauthoryear{Gal{\'a}rraga, Teflioudi, Hose, and
  Suchanek}{Gal{\'a}rraga et~al\mbox{.}}{2013}]%
        {galarraga2013amie}
\bibfield{author}{\bibinfo{person}{Luis~Antonio Gal{\'a}rraga},
  \bibinfo{person}{Christina Teflioudi}, \bibinfo{person}{Katja Hose}, {and}
  \bibinfo{person}{Fabian Suchanek}.} \bibinfo{year}{2013}\natexlab{}.
\newblock \showarticletitle{AMIE: association rule mining under incomplete
  evidence in ontological knowledge bases}. In
  \bibinfo{booktitle}{\emph{Proceedings of the 22nd international conference on
  World Wide Web}}. \bibinfo{pages}{413--422}.
\newblock


\bibitem[\protect\citeauthoryear{Gardner and Mitchell}{Gardner and
  Mitchell}{2015}]%
        {gardner2015efficient}
\bibfield{author}{\bibinfo{person}{Matt Gardner} {and} \bibinfo{person}{Tom
  Mitchell}.} \bibinfo{year}{2015}\natexlab{}.
\newblock \showarticletitle{Efficient and expressive knowledge base completion
  using subgraph feature extraction}. In \bibinfo{booktitle}{\emph{Proceedings
  of the 2015 Conference on Empirical Methods in Natural Language Processing}}.
  \bibinfo{pages}{1488--1498}.
\newblock


\bibitem[\protect\citeauthoryear{Grover and Leskovec}{Grover and
  Leskovec}{2016}]%
        {grover2016node2vec}
\bibfield{author}{\bibinfo{person}{Aditya Grover} {and} \bibinfo{person}{Jure
  Leskovec}.} \bibinfo{year}{2016}\natexlab{}.
\newblock \showarticletitle{node2vec: Scalable feature learning for networks}.
  In \bibinfo{booktitle}{\emph{Proceedings of the 22nd ACM SIGKDD International
  Conference on Knowledge Discovery and Data Mining (KDD)}}.
  \bibinfo{pages}{855--864}.
\newblock


\bibitem[\protect\citeauthoryear{Gu, Chang, Zhu, Sojoudi, and El~Ghaoui}{Gu
  et~al\mbox{.}}{2020}]%
        {gu2020implicit}
\bibfield{author}{\bibinfo{person}{Fangda Gu}, \bibinfo{person}{Heng Chang},
  \bibinfo{person}{Wenwu Zhu}, \bibinfo{person}{Somayeh Sojoudi}, {and}
  \bibinfo{person}{Laurent El~Ghaoui}.} \bibinfo{year}{2020}\natexlab{}.
\newblock \showarticletitle{Implicit graph neural networks}.
\newblock \bibinfo{journal}{\emph{Advances in Neural Information Processing
  Systems (NeurIPS)}}  \bibinfo{volume}{33} (\bibinfo{year}{2020}),
  \bibinfo{pages}{11984--11995}.
\newblock


\bibitem[\protect\citeauthoryear{Guan, Zhang, Li, Chang, Zhang, Qin, Jiang,
  Wang, and Zhu}{Guan et~al\mbox{.}}{2021}]%
        {guan2021autogl}
\bibfield{author}{\bibinfo{person}{Chaoyu Guan}, \bibinfo{person}{Ziwei Zhang},
  \bibinfo{person}{Haoyang Li}, \bibinfo{person}{Heng Chang},
  \bibinfo{person}{Zeyang Zhang}, \bibinfo{person}{Yijian Qin},
  \bibinfo{person}{Jiyan Jiang}, \bibinfo{person}{Xin Wang}, {and}
  \bibinfo{person}{Wenwu Zhu}.} \bibinfo{year}{2021}\natexlab{}.
\newblock \showarticletitle{AutoGL: A Library for Automated Graph Learning}. In
  \bibinfo{booktitle}{\emph{ICLR 2021 Workshop on Geometrical and Topological
  Representation Learning}}.
\newblock


\bibitem[\protect\citeauthoryear{Ho, Stepanova, Gad-Elrab, Kharlamov, and
  Weikum}{Ho et~al\mbox{.}}{2018}]%
        {ho2018rule}
\bibfield{author}{\bibinfo{person}{Vinh~Thinh Ho}, \bibinfo{person}{Daria
  Stepanova}, \bibinfo{person}{Mohamed~H Gad-Elrab}, \bibinfo{person}{Evgeny
  Kharlamov}, {and} \bibinfo{person}{Gerhard Weikum}.}
  \bibinfo{year}{2018}\natexlab{}.
\newblock \showarticletitle{Rule learning from knowledge graphs guided by
  embedding models}. In \bibinfo{booktitle}{\emph{International Semantic Web
  Conference}}. Springer, \bibinfo{pages}{72--90}.
\newblock


\bibitem[\protect\citeauthoryear{Huang, Li, Jiang, Cao, Lu, Yin, Subbian, Sun,
  and Wang}{Huang et~al\mbox{.}}{2022}]%
        {huang2022multilingual}
\bibfield{author}{\bibinfo{person}{Zijie Huang}, \bibinfo{person}{Zheng Li},
  \bibinfo{person}{Haoming Jiang}, \bibinfo{person}{Tianyu Cao},
  \bibinfo{person}{Hanqing Lu}, \bibinfo{person}{Bing Yin},
  \bibinfo{person}{Karthik Subbian}, \bibinfo{person}{Yizhou Sun}, {and}
  \bibinfo{person}{Wei Wang}.} \bibinfo{year}{2022}\natexlab{}.
\newblock \showarticletitle{Multilingual Knowledge Graph Completion with
  Self-Supervised Adaptive Graph Alignment}.
\newblock \bibinfo{journal}{\emph{arXiv preprint arXiv:2203.14987}}
  (\bibinfo{year}{2022}).
\newblock


\bibitem[\protect\citeauthoryear{Ji, Pan, Cambria, Marttinen, and Philip}{Ji
  et~al\mbox{.}}{2021}]%
        {ji2021survey}
\bibfield{author}{\bibinfo{person}{Shaoxiong Ji}, \bibinfo{person}{Shirui Pan},
  \bibinfo{person}{Erik Cambria}, \bibinfo{person}{Pekka Marttinen}, {and}
  \bibinfo{person}{S~Yu Philip}.} \bibinfo{year}{2021}\natexlab{}.
\newblock \showarticletitle{A survey on knowledge graphs: Representation,
  acquisition, and applications}.
\newblock \bibinfo{journal}{\emph{IEEE Transactions on Neural Networks and
  Learning Systems}} \bibinfo{volume}{33}, \bibinfo{number}{2}
  (\bibinfo{year}{2021}), \bibinfo{pages}{494--514}.
\newblock


\bibitem[\protect\citeauthoryear{Jin, Chang, Zhu, and Sojoudi}{Jin
  et~al\mbox{.}}{2021}]%
        {jin2021power}
\bibfield{author}{\bibinfo{person}{Ming Jin}, \bibinfo{person}{Heng Chang},
  \bibinfo{person}{Wenwu Zhu}, {and} \bibinfo{person}{Somayeh Sojoudi}.}
  \bibinfo{year}{2021}\natexlab{}.
\newblock \showarticletitle{Power up! Robust Graph Convolutional Network via
  Graph Powering}. In \bibinfo{booktitle}{\emph{Proceedings of the AAAI
  Conference on Artificial Intelligence (AAAI)}}, Vol.~\bibinfo{volume}{35}.
  \bibinfo{pages}{8004--8012}.
\newblock


\bibitem[\protect\citeauthoryear{Johansson, Shalit, and Sontag}{Johansson
  et~al\mbox{.}}{2016}]%
        {johansson2016learning}
\bibfield{author}{\bibinfo{person}{Fredrik Johansson}, \bibinfo{person}{Uri
  Shalit}, {and} \bibinfo{person}{David Sontag}.}
  \bibinfo{year}{2016}\natexlab{}.
\newblock \showarticletitle{Learning representations for counterfactual
  inference}. In \bibinfo{booktitle}{\emph{International conference on machine
  learning}}. \bibinfo{pages}{3020--3029}.
\newblock


\bibitem[\protect\citeauthoryear{Jung, Jung, and Kang}{Jung
  et~al\mbox{.}}{2020}]%
        {jung2020t}
\bibfield{author}{\bibinfo{person}{Jaehun Jung}, \bibinfo{person}{Jinhong
  Jung}, {and} \bibinfo{person}{U Kang}.} \bibinfo{year}{2020}\natexlab{}.
\newblock \showarticletitle{T-gap: Learning to walk across time for temporal
  knowledge graph completion}.
\newblock \bibinfo{journal}{\emph{arXiv preprint arXiv:2012.10595}}
  (\bibinfo{year}{2020}).
\newblock


\bibitem[\protect\citeauthoryear{Lao and Cohen}{Lao and Cohen}{2010}]%
        {lao2010relational}
\bibfield{author}{\bibinfo{person}{Ni Lao} {and} \bibinfo{person}{William~W
  Cohen}.} \bibinfo{year}{2010}\natexlab{}.
\newblock \showarticletitle{Relational retrieval using a combination of
  path-constrained random walks}.
\newblock \bibinfo{journal}{\emph{Machine learning}} \bibinfo{volume}{81},
  \bibinfo{number}{1} (\bibinfo{year}{2010}), \bibinfo{pages}{53--67}.
\newblock


\bibitem[\protect\citeauthoryear{Li, Han, Cheng, Su, Wang, Zhang, and Pan}{Li
  et~al\mbox{.}}{2019}]%
        {li2019predicting}
\bibfield{author}{\bibinfo{person}{Jia Li}, \bibinfo{person}{Zhichao Han},
  \bibinfo{person}{Hong Cheng}, \bibinfo{person}{Jiao Su},
  \bibinfo{person}{Pengyun Wang}, \bibinfo{person}{Jianfeng Zhang}, {and}
  \bibinfo{person}{Lujia Pan}.} \bibinfo{year}{2019}\natexlab{}.
\newblock \showarticletitle{Predicting path failure in time-evolving graphs}.
  In \bibinfo{booktitle}{\emph{Proceedings of the 25th ACM SIGKDD international
  conference on knowledge discovery \& data mining}}.
  \bibinfo{pages}{1279--1289}.
\newblock


\bibitem[\protect\citeauthoryear{Li, Huang, Chang, and Rong}{Li
  et~al\mbox{.}}{2022}]%
        {li2022semi}
\bibfield{author}{\bibinfo{person}{Jia Li}, \bibinfo{person}{Yongfeng Huang},
  \bibinfo{person}{Heng Chang}, {and} \bibinfo{person}{Yu Rong}.}
  \bibinfo{year}{2022}\natexlab{}.
\newblock \showarticletitle{Semi-Supervised Hierarchical Graph Classification}.
\newblock \bibinfo{journal}{\emph{IEEE Transactions on Pattern Analysis and
  Machine Intelligence}} (\bibinfo{year}{2022}).
\newblock


\bibitem[\protect\citeauthoryear{Lin, Lan, and Li}{Lin et~al\mbox{.}}{2021}]%
        {lin2021generative}
\bibfield{author}{\bibinfo{person}{Wanyu Lin}, \bibinfo{person}{Hao Lan}, {and}
  \bibinfo{person}{Baochun Li}.} \bibinfo{year}{2021}\natexlab{}.
\newblock \showarticletitle{Generative causal explanations for graph neural
  networks}. In \bibinfo{booktitle}{\emph{International Conference on Machine
  Learning}}. PMLR, \bibinfo{pages}{6666--6679}.
\newblock


\bibitem[\protect\citeauthoryear{Ma, Wan, Yang, Li, Hecht, and Teevan}{Ma
  et~al\mbox{.}}{2022}]%
        {ma2022learning}
\bibfield{author}{\bibinfo{person}{Jing Ma}, \bibinfo{person}{Mengting Wan},
  \bibinfo{person}{Longqi Yang}, \bibinfo{person}{Jundong Li},
  \bibinfo{person}{Brent Hecht}, {and} \bibinfo{person}{Jaime Teevan}.}
  \bibinfo{year}{2022}\natexlab{}.
\newblock \showarticletitle{Learning Causal Effects on Hypergraphs}. In
  \bibinfo{booktitle}{\emph{Proceedings of the 28th ACM SIGKDD Conference on
  Knowledge Discovery and Data Mining}}. \bibinfo{pages}{1202--1212}.
\newblock


\bibitem[\protect\citeauthoryear{Malliaros, Giatsidis, Papadopoulos, and
  Vazirgiannis}{Malliaros et~al\mbox{.}}{2020}]%
        {malliaros2020core}
\bibfield{author}{\bibinfo{person}{Fragkiskos~D Malliaros},
  \bibinfo{person}{Christos Giatsidis}, \bibinfo{person}{Apostolos~N
  Papadopoulos}, {and} \bibinfo{person}{Michalis Vazirgiannis}.}
  \bibinfo{year}{2020}\natexlab{}.
\newblock \showarticletitle{The core decomposition of networks: Theory,
  algorithms and applications}.
\newblock \bibinfo{journal}{\emph{The VLDB Journal}} \bibinfo{volume}{29},
  \bibinfo{number}{1} (\bibinfo{year}{2020}), \bibinfo{pages}{61--92}.
\newblock


\bibitem[\protect\citeauthoryear{Mansour, Mohri, and Rostamizadeh}{Mansour
  et~al\mbox{.}}{2009}]%
        {mansour2009domain}
\bibfield{author}{\bibinfo{person}{Yishay Mansour}, \bibinfo{person}{Mehryar
  Mohri}, {and} \bibinfo{person}{Afshin Rostamizadeh}.}
  \bibinfo{year}{2009}\natexlab{}.
\newblock \showarticletitle{Domain adaptation: Learning bounds and algorithms}.
\newblock \bibinfo{journal}{\emph{arXiv preprint arXiv:0902.3430}}
  (\bibinfo{year}{2009}).
\newblock


\bibitem[\protect\citeauthoryear{Morgan and Winship}{Morgan and
  Winship}{2015}]%
        {morgan2015counterfactuals}
\bibfield{author}{\bibinfo{person}{Stephen~L Morgan} {and}
  \bibinfo{person}{Christopher Winship}.} \bibinfo{year}{2015}\natexlab{}.
\newblock \bibinfo{booktitle}{\emph{Counterfactuals and causal inference}}.
\newblock \bibinfo{publisher}{Cambridge University Press}.
\newblock


\bibitem[\protect\citeauthoryear{Sadeghian, Armandpour, Ding, and
  Wang}{Sadeghian et~al\mbox{.}}{2019}]%
        {sadeghian2019drum}
\bibfield{author}{\bibinfo{person}{Ali Sadeghian},
  \bibinfo{person}{Mohammadreza Armandpour}, \bibinfo{person}{Patrick Ding},
  {and} \bibinfo{person}{Daisy~Zhe Wang}.} \bibinfo{year}{2019}\natexlab{}.
\newblock \showarticletitle{DRUM: End-To-End Differentiable Rule Mining On
  Knowledge Graphs}.
\newblock \bibinfo{journal}{\emph{Advances in Neural Information Processing
  Systems}}  \bibinfo{volume}{32}, \bibinfo{pages}{15347--15357}.
\newblock


\bibitem[\protect\citeauthoryear{Schlichtkrull, Kipf, Bloem, Van Den~Berg,
  Titov, and Welling}{Schlichtkrull et~al\mbox{.}}{2018}]%
        {schlichtkrull2018modeling}
\bibfield{author}{\bibinfo{person}{Michael Schlichtkrull},
  \bibinfo{person}{Thomas~N Kipf}, \bibinfo{person}{Peter Bloem},
  \bibinfo{person}{Rianne Van Den~Berg}, \bibinfo{person}{Ivan Titov}, {and}
  \bibinfo{person}{Max Welling}.} \bibinfo{year}{2018}\natexlab{}.
\newblock \showarticletitle{Modeling relational data with graph convolutional
  networks}. In \bibinfo{booktitle}{\emph{European semantic web conference}}.
  Springer, \bibinfo{pages}{593--607}.
\newblock


\bibitem[\protect\citeauthoryear{Shang, Tang, Huang, Bi, He, and Zhou}{Shang
  et~al\mbox{.}}{2019}]%
        {shang2019end}
\bibfield{author}{\bibinfo{person}{Chao Shang}, \bibinfo{person}{Yun Tang},
  \bibinfo{person}{Jing Huang}, \bibinfo{person}{Jinbo Bi},
  \bibinfo{person}{Xiaodong He}, {and} \bibinfo{person}{Bowen Zhou}.}
  \bibinfo{year}{2019}\natexlab{}.
\newblock \showarticletitle{End-to-end structure-aware convolutional networks
  for knowledge base completion}. In \bibinfo{booktitle}{\emph{Proceedings of
  the AAAI Conference on Artificial Intelligence}}, Vol.~\bibinfo{volume}{33}.
  \bibinfo{pages}{3060--3067}.
\newblock


\bibitem[\protect\citeauthoryear{Sun, Deng, Nie, and Tang}{Sun
  et~al\mbox{.}}{2019}]%
        {sun2019rotate}
\bibfield{author}{\bibinfo{person}{Zhiqing Sun}, \bibinfo{person}{Zhi-Hong
  Deng}, \bibinfo{person}{Jian-Yun Nie}, {and} \bibinfo{person}{Jian Tang}.}
  \bibinfo{year}{2019}\natexlab{}.
\newblock \showarticletitle{RotatE: Knowledge Graph Embedding by Relational
  Rotation in Complex Space}. In \bibinfo{booktitle}{\emph{International
  Conference on Learning Representations}}.
\newblock


\bibitem[\protect\citeauthoryear{Tang, Li, Gao, and Li}{Tang
  et~al\mbox{.}}{2022a}]%
        {tang2022rethinking}
\bibfield{author}{\bibinfo{person}{Jianheng Tang}, \bibinfo{person}{Jiajin Li},
  \bibinfo{person}{Ziqi Gao}, {and} \bibinfo{person}{Jia Li}.}
  \bibinfo{year}{2022}\natexlab{a}.
\newblock \showarticletitle{Rethinking Graph Neural Networks for Anomaly
  Detection}. In \bibinfo{booktitle}{\emph{International Conference on Machine
  Learning}}.
\newblock


\bibitem[\protect\citeauthoryear{{Tang}, {Qu}, {Wang}, {Zhang}, {Yan}, and
  {Mei}}{{Tang} et~al\mbox{.}}{2015}]%
        {WWW2015Line}
\bibfield{author}{\bibinfo{person}{Jian {Tang}}, \bibinfo{person}{Meng {Qu}},
  \bibinfo{person}{Mingzhe {Wang}}, \bibinfo{person}{Ming {Zhang}},
  \bibinfo{person}{Jun {Yan}}, {and} \bibinfo{person}{Qiaozhu {Mei}}.}
  \bibinfo{year}{2015}\natexlab{}.
\newblock \showarticletitle{LINE: Large-scale Information Network Embedding}.
  In \bibinfo{booktitle}{\emph{Proceedings of the 24th International Conference
  on World Wide Web}}. \bibinfo{pages}{1067--1077}.
\newblock


\bibitem[\protect\citeauthoryear{Tang, Pei, Zhang, Zhu, Zhuang, Hoehndorf, and
  Zhang}{Tang et~al\mbox{.}}{2022b}]%
        {tang2022positive}
\bibfield{author}{\bibinfo{person}{Zhenwei Tang}, \bibinfo{person}{Shichao
  Pei}, \bibinfo{person}{Zhao Zhang}, \bibinfo{person}{Yongchun Zhu},
  \bibinfo{person}{Fuzhen Zhuang}, \bibinfo{person}{Robert Hoehndorf}, {and}
  \bibinfo{person}{Xiangliang Zhang}.} \bibinfo{year}{2022}\natexlab{b}.
\newblock \showarticletitle{Positive-Unlabeled Learning with Adversarial Data
  Augmentation for Knowledge Graph Completion}.
\newblock \bibinfo{journal}{\emph{arXiv preprint arXiv:2205.00904}}
  (\bibinfo{year}{2022}).
\newblock


\bibitem[\protect\citeauthoryear{Toutanova and Chen}{Toutanova and
  Chen}{2015}]%
        {toutanova2015observed}
\bibfield{author}{\bibinfo{person}{Kristina Toutanova} {and}
  \bibinfo{person}{Danqi Chen}.} \bibinfo{year}{2015}\natexlab{}.
\newblock \showarticletitle{Observed versus latent features for knowledge base
  and text inference}. In \bibinfo{booktitle}{\emph{Proceedings of the 3rd
  workshop on continuous vector space models and their compositionality}}.
  \bibinfo{pages}{57--66}.
\newblock


\bibitem[\protect\citeauthoryear{Trouillon, Welbl, Riedel, Gaussier, and
  Bouchard}{Trouillon et~al\mbox{.}}{2016}]%
        {trouillon2016complex}
\bibfield{author}{\bibinfo{person}{Th{\'e}o Trouillon},
  \bibinfo{person}{Johannes Welbl}, \bibinfo{person}{Sebastian Riedel},
  \bibinfo{person}{{\'E}ric Gaussier}, {and} \bibinfo{person}{Guillaume
  Bouchard}.} \bibinfo{year}{2016}\natexlab{}.
\newblock \showarticletitle{Complex embeddings for simple link prediction}. In
  \bibinfo{booktitle}{\emph{International Conference on Machine Learning}}.
  PMLR, \bibinfo{pages}{2071--2080}.
\newblock


\bibitem[\protect\citeauthoryear{Van~der Laan and Petersen}{Van~der Laan and
  Petersen}{2007}]%
        {van2007causal}
\bibfield{author}{\bibinfo{person}{Mark~J Van~der Laan} {and}
  \bibinfo{person}{Maya~L Petersen}.} \bibinfo{year}{2007}\natexlab{}.
\newblock \showarticletitle{Causal effect models for realistic individualized
  treatment and intention to treat rules}.
\newblock \bibinfo{journal}{\emph{The international journal of biostatistics}}
  \bibinfo{volume}{3}, \bibinfo{number}{1} (\bibinfo{year}{2007}).
\newblock


\bibitem[\protect\citeauthoryear{Vashishth, Sanyal, Nitin, and
  Talukdar}{Vashishth et~al\mbox{.}}{2019}]%
        {vashishth2019composition}
\bibfield{author}{\bibinfo{person}{Shikhar Vashishth}, \bibinfo{person}{Soumya
  Sanyal}, \bibinfo{person}{Vikram Nitin}, {and} \bibinfo{person}{Partha
  Talukdar}.} \bibinfo{year}{2019}\natexlab{}.
\newblock \showarticletitle{Composition-based Multi-Relational Graph
  Convolutional Networks}. In \bibinfo{booktitle}{\emph{International
  Conference on Learning Representations}}.
\newblock


\bibitem[\protect\citeauthoryear{Wang, Zhou, Pan, Dong, Song, and Sha}{Wang
  et~al\mbox{.}}{2022}]%
        {Wang_Zhou_Pan_Dong_Song_Sha_2022}
\bibfield{author}{\bibinfo{person}{Changjian Wang}, \bibinfo{person}{Xiaofei
  Zhou}, \bibinfo{person}{Shirui Pan}, \bibinfo{person}{Linhua Dong},
  \bibinfo{person}{Zeliang Song}, {and} \bibinfo{person}{Ying Sha}.}
  \bibinfo{year}{2022}\natexlab{}.
\newblock \showarticletitle{Exploring Relational Semantics for Inductive
  Knowledge Graph Completion}.
\newblock \bibinfo{journal}{\emph{Proceedings of the AAAI Conference on
  Artificial Intelligence}} \bibinfo{volume}{36}, \bibinfo{number}{4}
  (\bibinfo{date}{Jun.} \bibinfo{year}{2022}), \bibinfo{pages}{4184--4192}.
\newblock
\urldef\tempurl%
\url{https://doi.org/10.1609/aaai.v36i4.20337}
\showDOI{\tempurl}


\bibitem[\protect\citeauthoryear{Wang, Zheng, Ye, Gan, Li, Song, Zhou, Ma, Yu,
  Gai, Xiao, He, Karypis, Li, and Zhang}{Wang et~al\mbox{.}}{2019}]%
        {dgl}
\bibfield{author}{\bibinfo{person}{Minjie Wang}, \bibinfo{person}{Da Zheng},
  \bibinfo{person}{Zihao Ye}, \bibinfo{person}{Quan Gan},
  \bibinfo{person}{Mufei Li}, \bibinfo{person}{Xiang Song},
  \bibinfo{person}{Jinjing Zhou}, \bibinfo{person}{Chao Ma},
  \bibinfo{person}{Lingfan Yu}, \bibinfo{person}{Yu Gai},
  \bibinfo{person}{Tianjun Xiao}, \bibinfo{person}{Tong He},
  \bibinfo{person}{George Karypis}, \bibinfo{person}{Jinyang Li}, {and}
  \bibinfo{person}{Zheng Zhang}.} \bibinfo{year}{2019}\natexlab{}.
\newblock \showarticletitle{Deep Graph Library: A Graph-Centric,
  Highly-Performant Package for Graph Neural Networks}.
\newblock  (\bibinfo{year}{2019}).
\newblock


\bibitem[\protect\citeauthoryear{Weiss, Kuusisto, Boyd, Liu, and Page}{Weiss
  et~al\mbox{.}}{2015}]%
        {weiss2015machine}
\bibfield{author}{\bibinfo{person}{Jeremy Weiss}, \bibinfo{person}{Finn
  Kuusisto}, \bibinfo{person}{Kendrick Boyd}, \bibinfo{person}{Jie Liu}, {and}
  \bibinfo{person}{David Page}.} \bibinfo{year}{2015}\natexlab{}.
\newblock \showarticletitle{Machine learning for treatment assignment:
  Improving individualized risk attribution}. In \bibinfo{booktitle}{\emph{AMIA
  Annual Symposium Proceedings}}, Vol.~\bibinfo{volume}{2015}. American Medical
  Informatics Association, \bibinfo{pages}{1306}.
\newblock


\bibitem[\protect\citeauthoryear{Yang, Yih, He, Gao, and Deng}{Yang
  et~al\mbox{.}}{2015}]%
        {yang2015embedding}
\bibfield{author}{\bibinfo{person}{Bishan Yang}, \bibinfo{person}{Wen-tau Yih},
  \bibinfo{person}{Xiaodong He}, \bibinfo{person}{Jianfeng Gao}, {and}
  \bibinfo{person}{Li Deng}.} \bibinfo{year}{2015}\natexlab{}.
\newblock \showarticletitle{Embedding entities and relations for learning and
  inference in knowledge bases}. In \bibinfo{booktitle}{\emph{International
  Conference on Learning Representations}}.
\newblock


\bibitem[\protect\citeauthoryear{Yang, Yang, and Cohen}{Yang
  et~al\mbox{.}}{2017}]%
        {yang2017differentiable}
\bibfield{author}{\bibinfo{person}{Fan Yang}, \bibinfo{person}{Zhilin Yang},
  {and} \bibinfo{person}{William~W Cohen}.} \bibinfo{year}{2017}\natexlab{}.
\newblock \showarticletitle{Differentiable learning of logical rules for
  knowledge base reasoning}. In \bibinfo{booktitle}{\emph{Advances in Neural
  Information Processing Systems}}. \bibinfo{pages}{2316--2325}.
\newblock


\bibitem[\protect\citeauthoryear{Yao, Chu, Li, Li, Gao, and Zhang}{Yao
  et~al\mbox{.}}{2020}]%
        {yao2020causal}
\bibfield{author}{\bibinfo{person}{Liuyi Yao}, \bibinfo{person}{Zhixuan Chu},
  \bibinfo{person}{Sheng Li}, \bibinfo{person}{Yaliang Li},
  \bibinfo{person}{Jing Gao}, {and} \bibinfo{person}{Aidong Zhang}.}
  \bibinfo{year}{2020}\natexlab{}.
\newblock \bibinfo{title}{A Survey on Causal Inference}.
\newblock
\newblock
\urldef\tempurl%
\url{https://doi.org/10.48550/ARXIV.2002.02770}
\showDOI{\tempurl}


\bibitem[\protect\citeauthoryear{Zhang, Cai, Zhang, and Wang}{Zhang
  et~al\mbox{.}}{2020a}]%
        {zhang2020learning}
\bibfield{author}{\bibinfo{person}{Zhanqiu Zhang}, \bibinfo{person}{Jianyu
  Cai}, \bibinfo{person}{Yongdong Zhang}, {and} \bibinfo{person}{Jie Wang}.}
  \bibinfo{year}{2020}\natexlab{a}.
\newblock \showarticletitle{Learning hierarchy-aware knowledge graph embeddings
  for link prediction}. In \bibinfo{booktitle}{\emph{Proceedings of the AAAI
  Conference on Artificial Intelligence}}, Vol.~\bibinfo{volume}{34}.
  \bibinfo{pages}{3065--3072}.
\newblock


\bibitem[\protect\citeauthoryear{Zhang, Cui, and Zhu}{Zhang
  et~al\mbox{.}}{2020b}]%
        {zhang2020deep}
\bibfield{author}{\bibinfo{person}{Ziwei Zhang}, \bibinfo{person}{Peng Cui},
  {and} \bibinfo{person}{Wenwu Zhu}.} \bibinfo{year}{2020}\natexlab{b}.
\newblock \showarticletitle{Deep learning on graphs: A survey}.
\newblock \bibinfo{journal}{\emph{IEEE Transactions on Knowledge and Data
  Engineering (TKDE)}} (\bibinfo{year}{2020}).
\newblock


\bibitem[\protect\citeauthoryear{Zhang, Wang, Ye, and Wu}{Zhang
  et~al\mbox{.}}{2022}]%
        {zhang2022rethinking}
\bibfield{author}{\bibinfo{person}{Zhanqiu Zhang}, \bibinfo{person}{Jie Wang},
  \bibinfo{person}{Jieping Ye}, {and} \bibinfo{person}{Feng Wu}.}
  \bibinfo{year}{2022}\natexlab{}.
\newblock \showarticletitle{Rethinking Graph Convolutional Networks in
  Knowledge Graph Completion}. In \bibinfo{booktitle}{\emph{Proceedings of the
  ACM Web Conference 2022}}. \bibinfo{pages}{798--807}.
\newblock


\bibitem[\protect\citeauthoryear{Zhao, Liu, Wang, Yu, and Jiang}{Zhao
  et~al\mbox{.}}{2021}]%
        {zhao2021counterfactual}
\bibfield{author}{\bibinfo{person}{Tong Zhao}, \bibinfo{person}{Gang Liu},
  \bibinfo{person}{Daheng Wang}, \bibinfo{person}{Wenhao Yu}, {and}
  \bibinfo{person}{Meng Jiang}.} \bibinfo{year}{2021}\natexlab{}.
\newblock \showarticletitle{Counterfactual graph learning for link prediction}.
\newblock \bibinfo{journal}{\emph{arXiv preprint arXiv:2106.02172}}
  (\bibinfo{year}{2021}).
\newblock


\bibitem[\protect\citeauthoryear{Zhao, Liu, Wang, Yu, and Jiang}{Zhao
  et~al\mbox{.}}{2022}]%
        {zhao2022learning}
\bibfield{author}{\bibinfo{person}{Tong Zhao}, \bibinfo{person}{Gang Liu},
  \bibinfo{person}{Daheng Wang}, \bibinfo{person}{Wenhao Yu}, {and}
  \bibinfo{person}{Meng Jiang}.} \bibinfo{year}{2022}\natexlab{}.
\newblock \showarticletitle{Learning from counterfactual links for link
  prediction}. In \bibinfo{booktitle}{\emph{International Conference on Machine
  Learning}}. PMLR, \bibinfo{pages}{26911--26926}.
\newblock


\bibitem[\protect\citeauthoryear{Zhu, Xu, Yu, Liu, Wu, and Wang}{Zhu
  et~al\mbox{.}}{2021a}]%
        {zhu2021graph}
\bibfield{author}{\bibinfo{person}{Yanqiao Zhu}, \bibinfo{person}{Yichen Xu},
  \bibinfo{person}{Feng Yu}, \bibinfo{person}{Qiang Liu}, \bibinfo{person}{Shu
  Wu}, {and} \bibinfo{person}{Liang Wang}.} \bibinfo{year}{2021}\natexlab{a}.
\newblock \showarticletitle{Graph contrastive learning with adaptive
  augmentation}. In \bibinfo{booktitle}{\emph{Proceedings of the Web Conference
  2021}}. \bibinfo{pages}{2069--2080}.
\newblock


\bibitem[\protect\citeauthoryear{Zhu, Zhang, Xhonneux, and Tang}{Zhu
  et~al\mbox{.}}{2021b}]%
        {zhu2021neural}
\bibfield{author}{\bibinfo{person}{Zhaocheng Zhu}, \bibinfo{person}{Zuobai
  Zhang}, \bibinfo{person}{Louis-Pascal Xhonneux}, {and} \bibinfo{person}{Jian
  Tang}.} \bibinfo{year}{2021}\natexlab{b}.
\newblock \showarticletitle{Neural bellman-ford networks: A general graph
  neural network framework for link prediction}.
\newblock \bibinfo{journal}{\emph{Advances in Neural Information Processing
  Systems}}  \bibinfo{volume}{34} (\bibinfo{year}{2021}).
\newblock


\end{thebibliography}

%%
%% If your work has an appendix, this is the place to put it.
\appendix

\section{Overall Algorithm of \method.}
Alg.~\ref{alg:method} summarizes the overall algorithm of \method during the training and inference stages. Note that only the output $p^{F}(t | h, r)$ from the factual treatments will be used during the inference stage for the task of KGC.

\begin{algorithm}[htbp]
   \caption{The overall algorithm of \method.}
   \label{alg:method}
\begin{algorithmic}
    \STATE {\bfseries Input:} Encoder $f$, decoder $g$, training triplets $\{(h, r, t)\}$, factual adjacency matrix $\gA^{F}$, number of training epochs $n\_epochs$
    \STATE Compute factual treatments $\gT^{F}$ as presented in Sec.~\ref{sec:TV}.
    \STATE Compute counterfactual $\gT^{CF}, \gA^{CF}$ by Eq.(\ref{eq:nearest2}) and (\ref{eq:tcf}).
    \STATE \verb+// model training+
    \FOR{epoch in range($n\_epochs$)}
        \STATE $\vz_r(h,t) = f\left((h,r,t)\right)$ with Eq.(\ref{eq:gcn_layer}).
        \STATE Get $p^{F}(t | h, r)$ and $p^{CF}(t | h, r)$ via $g$ with Eq.(\ref{eq:apredf}) and (\ref{eq:apredcf}).
        \STATE Update parameters $\Theta_f$ and $\Theta_g$ in $f$ and $g$, respectively, with $\mathcal{L}$ (\ref{eq:loss}).
    \ENDFOR
    \STATE \verb+// inference+
    \STATE $\vz_r(h,t) = f\left((h,r,t)\right)$ with Eq.(\ref{eq:gcn_layer}).
    \STATE Get $p^{F}(t | h, r)$ and $p^{CF}(t | h, r)$ via $g$ with Eq.(\ref{eq:apredf}) and (\ref{eq:apredcf}).
    \STATE {\bfseries Output:} $p^{F}(t | h, r)$ for the task of KGC, $p^{CF}(t | h, r)$.
\end{algorithmic}
\end{algorithm}

\begin{table}[htbp]
    \centering
    \caption{Dataset statistics for KGC.}
    \label{tab:kg_statistics}
    \resizebox{\columnwidth}{!}{%
    \begin{tabular}{lccccc}
    \toprule
    \multirow{2}{*}{\bf{Dataset}} & \multirow{2}{*}{\bf{\#Entity}} & \multirow{2}{*}{\bf{\#Relation}} & \multicolumn{3}{c}{\bf{\#Triplet}} \\
    & & & \bf{\#Train} & \bf{\#Validation} & \bf{\#Test} \\
    \midrule
    FB15k-237~\cite{toutanova2015observed} & 14,541 & 237 & 272,115 & 17,535 & 20,466 \\
    WN18RR~\cite{dettmers2018convolutional} & 40,943 & 11 & 86,835 & 3,034 & 3,134 \\
    \bottomrule
    \end{tabular}
    }
\end{table}

\section{Statistics of datasets}
Dataset statistics of FB15k-237 and WN18RR for KGC are summarized in Table~\ref{tab:kg_statistics}. We use the standard splits~\cite{toutanova2015observed,dettmers2018convolutional} for a fair comparison.

\section{Evaluation metrics}
We follow the standard filtered ranking protocol~\cite{bordes2013translating} and evaluate the performance of models by five metrics Mean Reciprocal Rank (MRR), Mean Rank (MR), and Hits@K (H@K). The detailed definitions of five metrics is introduced below:

\xhdr{MRR}
For a test triplet $(h, r, t)$, we rank it against all negative triplets $(h', r, t)$ or $(h, r, t')$ obtained by corrupting the head 
(or the tail) of the relation.
Let $\rank_h(h,r,t)$ be the ranking of $(h, r, t)$ among all head-corrupted relations, and 
let $\rank_t(h,r,t)$ denote a similar ranking with tail corruptions.
MRR is the mean of the reciprocal rank:
\begin{align*} %\label{eq:mrr}
\func{MRR} = \frac{1}{2 * |\set{T}|} \sum_{(h, r, t) \in \set{T}} \Big( \frac{1}{\rank_h(h,r,t)} + \frac{1}{\rank_t(h,r,t)} \Big),
\end{align*}
where $\set{T}$ is the test set.

\xhdr{MR}
MR is the mean of both ranks:
\begin{align*}
\func{MRR} = \frac{1}{2 * |\set{T}|} \sum_{(h, r, t) \in \set{T}} \Big(\rank_h(h,r,t) + \rank_t(h,r,t)\Big),
\end{align*}

\xhdr{Hits@K}
Hits@K measures the proportion of triples in $\set{T}$ that rank among top-K after corrupting both heads and tails. 

\end{document}